%% file: main_arxiv.tex
\documentclass{ieeetj}
\usepackage{amsmath,amssymb,amsfonts}
\usepackage{graphicx,color}
\usepackage{textcomp}
\usepackage{xcolor}
\usepackage[table]{xcolor}
\def\BibTeX{{\rm B\kern-.05em{\sc i\kern-.025em b}\kern-.08em
    T\kern-.1667em\lower.7ex\hbox{E}\kern-.125emX}}
\AtBeginDocument{\definecolor{tmlcncolor}{cmyk}{0.93,0.59,0.15,0.02}\definecolor{NavyBlue}{RGB}{0,86,125}}

\def\OJlogo{\vspace{-4pt}$<$Society logo(s) and publication title will appear here.$>$}
\def\seclogo{\vspace{10pt}$<$Society logo(s) and publication title will appear here.$>$}

\def\authorrefmark#1{\ensuremath{^{\textbf{#1}}}}

\input{preamble.tex}
\begin{document}
\receiveddate{XX Month, XXXX}
\reviseddate{XX Month, XXXX}
\accepteddate{XX Month, XXXX}
\publisheddate{XX Month, XXXX}
\currentdate{XX Month, XXXX}
\doiinfo{XXXX.2022.1234567}

\markboth{}{Boxan {et al.}}

\title{One year in a forest: Analyzing the challenges of autonomous navigation in subarctic environments}

\author{
Mat\v{e}j Boxan\authorrefmark{1},
Nicolas Lauzon\authorrefmark{1},
Veronica Vannini\authorrefmark{1},\\
Mathis Turgeon-Roy\authorrefmark{1}, and
François Pomerleau\authorrefmark{1} (Senior Member, IEEE)}
\affil{Norlab, Université Laval, Quebec, Canada}
\corresp{Corresponding author: Mat\v{e}j Boxan (email: matej.boxan@norlab.ulaval.ca).}
\authornote{
This research was supported by the Natural Sciences and Engineering Research Council of Canada (NSERC) and the Fonds de recherche du Québec (FRQNT) through the grant 2023-NOVA-326877 HUNTER (Highlight the Unexpected with Navigation Through Extreme Regions).
Additionally, this project benefited from a Canada Foundation for Innovation Fund grant (\#39709, PI: E. Thiffault and F. Anctil).
}

\begin{abstract}
Subarctic regions have the potential to see increased deployment of autonomous robots in applications including forestry, mining, and environmental monitoring.
In these conditions, an autonomous system's reliance on \ac{GNSS} or cloud computing is precarious due to dense tree canopies and atmospheric attenuation, necessitating onboard sensing and data processing.
However, established exteroceptive modalities, including cameras, lidars, and radars, are typically evaluated in structured urban settings or in environments that lack significant seasonal variations.
To address this, we present a field report on a year-long deployment of a mobile robot in a subarctic boreal forest.
We evaluate \qty{64}{\kilo\meter} of data using nine odometry, localization, and mapping methods, and assess their performance across seasonal changes that include high snow accumulation and temperature shifts of \qty{60}{\degreeCelsius}.
\rev{The performed experiments suggest that} the environment changes significantly hinder the performance of state-of-the-art techniques, which show increased fragility when subject to conditions characterized by self-similar scenes or tall snowbanks.
Additionally, complex \ac{SLAM} algorithms offer limited accuracy gains over a proprioceptive baseline while significantly increasing system fragility.
Furthermore, by correlating the position drift with features and confidence weight distribution, we show that visual-based \ac{SLAM} methods are particularly affected by the seasonal changes.
Additionally, we investigate the task of cross-season localization in a prior map.
While lidar-based methods successfully completed localization runs between seasons, radar and visual methods are prone to failure due to a few matching features between runs, even within the same season.
Finally, we detail the challenges and lessons learned from this year-long trial, including a multi-season \acf{TaR} evaluation using both radar and lidar-based~pipelines.
\end{abstract}



\maketitle

\acresetall

\section{INTRODUCTION}
Subarctic regions have been seeing increased attention in recent years.
Notably, while currently difficult to access, rapid climate-induced changes~\cite{Rees2020_climate} suggest these regions will see major human development in the near future~\cite{Stephani2022_infrastructure}.
Although these zones possess vast natural resources, their harsh nature, characterized by limited sunlight, prolonged winters, and the absence of basic infrastructure, severely constrains human activities.
Consequently, such conditions make the region ideal for the deployment of autonomous systems in sectors such as mining~\cite{Chen2024_mining}, forestry~\cite{Semberg2024_forwarder}, or inspection applications~\cite{Dandurand2022_inspection}.
However, for these autonomous systems to be viable, they must overcome the same environmental hostilities that have long sidelined human activity in the cryosphere~\cite{Pomerleau2023_snow}.
Autonomous operations in subarctic regions face severe technical constraints.
First, reliance on \ac{GNSS} is precarious.
Satellites appear low on the horizon, while snow-laden canopies in boreal forests obscure the line-of-sight~\cite{deJong2014_gnss_arctic, Brach2019_gnss_forest}.
Atmospheric attenuation from snow, rain, or humidity can absorb high-frequency signals, causing bandwidth drops~\cite{Ullah2025_satellite}, while ice accumulation on receivers further degrades performance~\cite{Amaya2014_satellite}.
Second, since satellite internet antennas face the same obstructions as \ac{GNSS} and cellular networks are limited, computing must be performed on the edge.
Power management presents another critical challenge.
Cold temperatures reduce battery capacity while the complex terrains, including mud and snow,  increase energy demand required for \acp{UGV} locomotion~\cite{Luo2022_batteries, Baril2022}.
Last but not least, systems must achieve an exceptional reliability, as maintenance in remote areas is both logistically complex and prohibitively expensive~\cite{Barabadi2011_reliability}.

\begin{figure}[thbp]
    \centering
    \includegraphics[width=\linewidth]{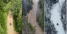}
    \caption{Example of seasonal changes in subarctic climate affecting a boreal forest.
    These conditions are used to evaluate the impact of seasonal variations on the performance of odometry, localization, and mapping algorithms.
    To this extent, we study the effects of seasonal changes on radar, lidar, and visual-based methods, based on a field campaign spanning one year.}
    \label{fig:intro}
\end{figure}

While the subarctic region encompasses a diverse mosaic of ecosystems, ranging from the alpine biome to permafrost tundra to shrublands, our focus is on the boreal forest, Earth's largest land biome~\cite{Hayes2022}.
Driven by rapid northern expansion, the biome has grown by over \qty{12}{\percent}, equivalent to \qty{0.84}{million~\kilo\meter\squared}, in the past four decades~\cite{Feng2026_northward}.
We specifically examine autonomous navigation in humid subarctic regions that witness major seasonal changes due to high snow accumulation and temperature swings.
An example of seasonal variations is illustrated in \autoref{fig:intro}, depicting a robot driving through the same location in summer, autumn and winter.
Changes in scene appearance present a substantial challenge for the deployment of autonomous systems, as standard localization and place recognition algorithms struggle when subject to high seasonal variations \cite{Schmidt2025_rover, Baril2022}.

In this work, we present an analysis of nine odometry, localization, and mapping techniques subject to perturbations caused by the subarctic environment.
\rev{Rather than isolating a single optimal method, this analysis is intended to highlight trends in how different approaches respond to seasonal variations.}
We base our analysis on a year-long deployment of a mobile robot in a typical exemplar of a boreal forest.
The environmental site is defined by major seasonal changes with a temperature difference between summer and winter of over \qty{60}{\degreeCelsius}, up to six meters of annual snowfall with snow accumulation of over one meter in open areas~\cite{Hess2023}.
We use our year-long deployment to define a framework for time series evaluation, clarifying the terminology and methods for the evaluation of autonomous navigation algorithms in the context of field campaigns that see high failure rates.
The main contributions of this work are:
\begin{itemize}
    \item A thorough analysis of the impact of the subarctic region on \rev{nine} state-of-the-art lidar, radar, and visual-based odometry techniques throughout the year;
    \item Characterization of the fragility of \ac{SLAM} systems in subarctic environments with respect to the complexity of the underlying algorithms;
    \item A study on cross-season localization in the subarctic region for two lidar-, \rev{two} visual-, and a radar-based method.
\end{itemize}

\section{RELATED WORK}

To position our scientific contributions, we start by reviewing the literature relevant to deploying autonomous systems in subarctic environments.
Existing literature typically divides environments for autonomous driving into simple on-road and off-road categories.
However, the wide variety of unstructured environments encountered during field deployments necessitates a finer classification of off-road conditions based on terrain complexity.
Consequently, we distinguish between off-pavement and off-trail terrain.
Off-pavement terrains feature visually and structurally distinct pathways, including gravel roads, forestry tracks, or firebreaks.
Conversely, off-trail terrain lacks any predefined path, forcing the vehicle to negotiate raw ground, vegetation, and natural obstacles.

To help the categorization of odometry, localization, and mapping across changing seasons, we will use the Köppen-Geiger Climate Classification~\cite{Peel2017}.
This system classifies regions based on seasonal temperature and precipitation thresholds, which directly correlate to the seasonal challenges faced by mobile robots.
A summary of the subset of Köppen-Geiger climate zones discussed in this paper is presented in~\autoref{tab:koppen_climates}.
Building on this terminology, we first examine the challenges to sensor modalities inherent in the multi-season conditions of wet subarctic climate zones.
Subsequently, we discuss the state of the art in localization and mapping across changing seasons.

\begin{table}[ht]
\centering
\caption{Köppen-Geiger climate classification for a subset of zones where robots were deployed over multiple seasons.
The Subarctic climate zone is the focus of this paper.}
\label{tab:koppen_climates}
\begin{tabularx}{\columnwidth}{@{}rlp{2.4cm}X@{}}
\toprule
& \textbf{Code} & \textbf{Climate Name} & \textbf{Criteria} \\
\midrule
\multirow{3}{*}[-2mm]{\rotatebox[origin=c]{90}{\emph{temperate}}}
& \textbf{Cfa} & Humid subtropical & Hot humid summer, mild winter \\
& \textbf{Cfb} & Temperate oceanic & Warm summer, mild winter \\
& \textbf{Csa} & Hot-summer Mediterranean & Hot dry summer, mild winter \\
\midrule
\multirow{4}{*}[-2mm]{\rotatebox[origin=c]{90}{\emph{continental}}}
& \textbf{Dfa} & Humid continental & Hot summer, 1 month above \qty{22}{\degreeCelsius} \\
& \textbf{Dfb} & Humid continental & Warm summer \\
& \textbf{Dfc} & \emph{Subarctic}         & Coldest month below \qty{-3}{\degreeCelsius} \\
& \textbf{Dwa} & Monsoon-influenced humid continental & Hot rainy summer, dry winter \\
\bottomrule
\end{tabularx}
\end{table}

\subsection{EFFECTS OF SEASONAL VARIATIONS ON AUTONOMOUS DRIVING SENSING}

Autonomous vehicles rely on sensor fusion from multiple modalities to navigate safely and efficiently through complex environments.
However, these sensor modalities are impacted by the changes brought by seasonal variations to a different degree.
All sensing, including \acp{IMU}, is subject to intrinsic drift caused by fluctuating temperature and humidity \cite{Yang2023_selfcalib}.
In certain climate zones, seasonal temperature differentials can exceed \qty{60}{\degreeCelsius}, rendering standard factory calibration insufficient~\cite{Martinez2022_mems}.
Furthermore, relying on external measurements also faces distinct seasonal challenges.
Tree canopies can obstruct \ac{GNSS} signal in the vegetative season, while winter conditions can lead to snow and ice accumulation on antennas, lowering the signal-to-noise ratio~\cite{Amaya2014_satellite, Kubelka2020_radio}.
Operational challenges also extend to the infrastructure required for autonomous navigation.
For example, static equipment, including \ac{RTK} \ac{GNSS} base antennas, risks sinking or shifting in snow over time~\cite{Baril2022}, necessitating robust, permanent structures to ensure stable operations.

Cameras are information-rich, but remain vulnerable to illumination dynamics.
While surface albedo is relatively uniform in summer months, winter introduces snow, which is highly reflective compared to other environmental features~\cite{Paton2017_expanding}.
Similarly, summer tree canopies create high-frequency illumination changes.
In these high contrast scenarios, standard auto-exposure algorithms fail due to either saturation or underexposure~\cite{Gamache2025}.
\rev{Additionally, rather than relying on explicit heuristics like the gray world assumption, deep learning methods are highly vulnerable to representation bias, which arises from how data is sampled during the collection process \cite{Mehrabi2021_survey}.}
Researchers typically mitigate the issue with data augmentation or domain adaptation during training~\cite{Lin2025_domain}.

Historically, research in range sensors has focused on adverse weather, such as fog, rain, or snowfall~\cite{Duthon2019_light, Kutila2018_automotive, Hong2020_RadarSLAM}.
\citet{Courcelle2023_importance} highlighted the danger of snow gusts, where a large volume of snow moves through the environment, possibly confusing localization and mapping algorithms.
Moreover, research shows that snow-covered objects are more difficult to detect by 3D~segmentation algorithms~\cite{Tang2025_cadc+}.
\rev{Looking under ground, \acp{GPR} have demonstrated robust long-term localization capabilities across changing seasons \cite{Ort2020_autonomous}.
However, they are susceptible to degradation because shifting soil water content alters local dielectric properties, introducing measurement inaccuracies.}
Current \ac{FMCW} rotating radars operate in two dimensions and are therefore strongly influenced by surface conditions.
Off-road, soft substrates, such as snow or mud, evolve with the seasons, further increasing the dynamic conditions present in the environments.
The inconsistent vehicle attitude introduces ground strikes, making feature matching more difficult~\cite{Qiao2025_thesis}.
Ultimately, while both lidars and radars perform adequately in low-to-moderate precipitation, the primary challenge stems from long-term, cross-season environmental structural changes, such as the formation of snowbanks~\cite{Baril2022}.
In this work, we extend the state of the art by reporting the impact of seasonal variations in subarctic environments on cameras, lidars and radars across seasons.

\subsection{LOCALIZATION AND MAPPING IN MULTI-SEASON CONDITIONS}

Recent advancements have enabled autonomous robots to execute complex tasks across a variety of environments.
While state-of-the-art systems demonstrate high performance in demanding conditions, including dynamic scenes, illumination variations, and adverse weather, the specific challenges posed by seasonal changes have not yet been the subject of broader research interest.
Indeed, a literature review by~\citet{Sousa2023} on long-term localization and mapping for mobile robots identifies only \qty{8}{\percent} of methods that explicitly address seasonal transitions.
Furthermore, most works evaluating the performance of \ac{SLAM}, odometry, and localization algorithms focus on urban environments.

These methods benefit from long-term data collections, such as Boreas~\cite{Burnett2023_boreas}, Oxford RobotCar~\cite{Maddern2017}, or the 4Seasons~\cite{Wenzel2021} datasets.
While such data recordings are crucial for both place recognition \cite{Uy2018_pointnetvlad, Kim2022} and lifelong mapping~\cite{Kim2022_ltmapper}, inter-session changes in urban environments usually come from dynamic objects, such as cars and pedestrians, or construction sites~\cite{Kim2022_ltmapper}.
Traditionally, map elements are modeled as either static or dynamic, an assumption that fails for semi-static features such as parked cars or seasonal snow accumulation.
\citet{Gil2025_ephemerality} addressed the challenge by modeling elements of the world into two-stage ephemerality, defining the transiency of points within two different time scales.
However, seasonal changes in off-road environments also modify the sensors' point of view, as the vehicle's elevation and attitude change with the season.
Furthermore, the effect of seasonal changes in urban environments is mitigated by human interventions, preventing the natural accumulation of snow or leaves.
Visual methods, such as the work of~\citet{Berrio2022, Bouaziz2023_solar}, profit by this constant upkeep, as the maintenance ensures clear roads for place recognition.
Additionally, urban settings provide artificial illumination from street lamps, surrounding traffic, and other ambient light, while static structures offer geometric features that remain stable across seasons.
In contrast, we analyze the effects of seasonal variations in a boreal forest.
Our work therefore includes test sequences including both limited human interventions, such as snowplowing, as well as entirely unmaintained conditions.

\begin{figure}[tbp]
    \centering
    \includegraphics[width=\linewidth]{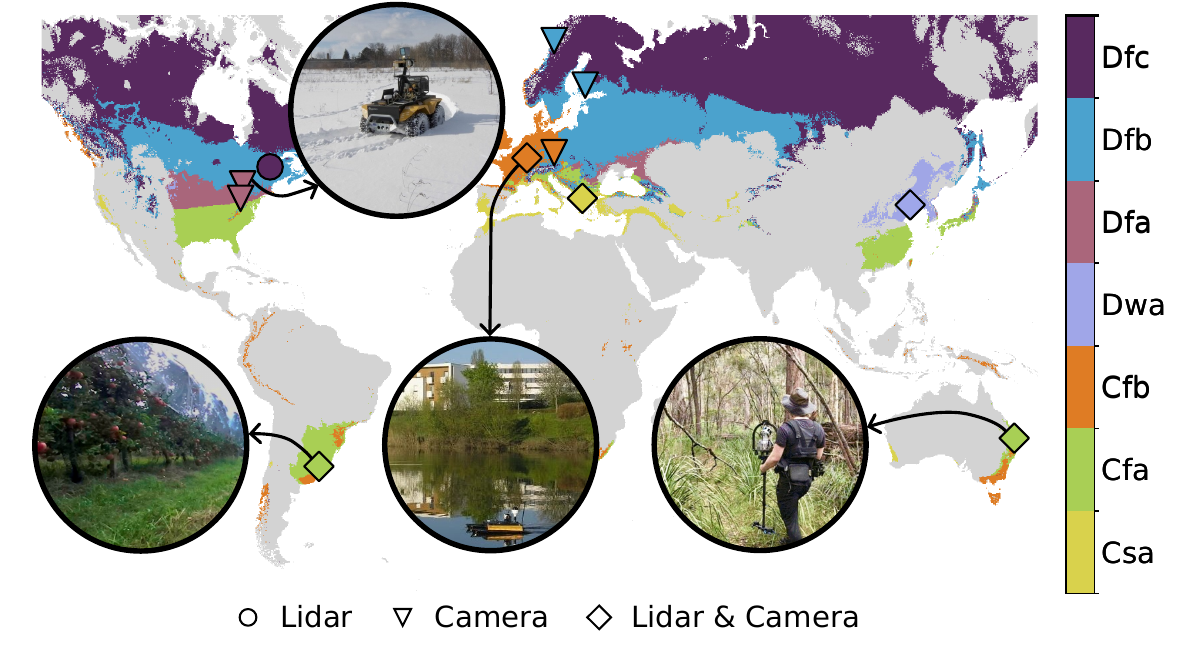}
    \caption{
        Distribution of data recordings and localization and mapping methods that include seasonal variations across the globe, with their Köppen-Geiger climate classification.
        For simplicity, we do not include climate zones not covered by multi-season datasets or not present in the evaluation of localization and mapping methods in unstructured environments.
        The map was cropped to focus on the identified deployments.
        Image inset were adapted from \cite{Marzoa2024, Paton2017_expanding, Griffith2017, Knights2023}.
    }
    \label{fig:related-work_map}
\end{figure}

Outside of maintained urban landscapes, the impact of seasonal changes on robot localization and mapping becomes significantly more pronounced.
Nevertheless, the severity of these changes varies by region, and the cyclic alternations are tightly linked to climate zones.
\autoref{fig:related-work_map} displays a map of geographical regions where multi-season mobile robot deployments have been conducted in unstructured environments.
The figure displays the deployment locations, together with their Köppen-Geiger climate classification.

In milder climates, challenges are primarily related to vegetation growth.
For example, the MAgro Dataset~\cite{Marzoa2024} captures eight months of sensor data recorded in an agricultural context in a humid subtropical (Cfa) climate zone in Uruguay.
Similarly, in the hot-summer Mediterranean (Csa) zone, the BLT dataset~\cite{Polvara2024} documents vineyard transformations between March and September.
\citet{Hroob2024} later utilized the BLT lidar data for long-term scan-to-map localization.
Moving to the oceanic (Cfb) zone, \rev{\citet{Krajnik2010_simple} showed that a \ac{TaR} system, relying on wheel odometry for translation and monocular camera for heading, could successfully navigate monthly seasonal appearance transitions, including snow cover, between October and March.}
The Symphony Lake Dataset by~\citet{Griffith2017} captures three years of visual data of a lake shore in Metz, France.
The seasonal variations in this dataset were dominated by foliage cycles of vegetation around the lake.
\citet{Griffith2019} subsequently demonstrated that a map-centric approach outperformed standard appearance-based methods in matching images across seasons.
Unlike milder climate zones, the studied subarctic environment introduces major volumetric transformations of the ground plane caused by snow accumulation, altering the vehicle's attitude and sensor perspective.

Continental climate zones experience intensified seasonal contrasts due to significant annual temperature variations.
\citet{Paton2017_expanding} tested a multi-channel Visual \ac{TaR} system, qualifying the influence of the diurnal cycle on localization across seasons in the hot summer humid continental (Dfa) climate zone.
\citet{MacTavish2018} extended their work with a multi-experience visual localization system over a 100-day-long deployment.
Further north, in the warm-summer humid continental (Dfb) climate, the FinnForest dataset~\cite{Ali2020} captures distinct summer and winter conditions in a Finnish coniferous forest.
\citet{Pritchard2025} leveraged the data for a robust forest-specific visual odometry system.
Additionally, the Nordlandsbanen dataset\footnote{\url{https://nrkbeta.no/2013/01/15/nordlandsbanen-minute-by-minute-season-by-season}} spans all four seasons across over \qty{700}{\kilo\meter} train line in Norway, crossing from the humid continental (Dfb) into the subarctic (Dfc) zone.
Recent efforts in the subarctic zone include the work of~\citet{Baril2022}, who evaluated \ac{TaR} performance between winter and autumn.
We extend this work by evaluating a wider range of sensor modalities, including lidar, radar, and cameras, across the full annual cycle.

Beyond localization and mapping, seasonal changes pose a major challenge for place recognition algorithms, a critical component for loop closure and re-localization in many \ac{SLAM} pipelines.
Among the primary studies in this area, \citet{Sattler2018} established the first benchmark for visual place recognition incorporating seasonal variations.
Their CMU-seasons dataset, recorded in the humid subtropical (Dfa) climate zone, was later used, alongside the Symphony Lake dataset, to develop an image-based place recognition algorithm targeting natural environments subject to seasonal changes~\cite{Benbihi2020}.
\rev{\citet{Neubert2015_superpixel} tested their predictive place recognition method relying on vocabularies of superpixels against the Nordlandsbanen dataset (Dfb, Dfc).
Other researchers used the same data recording to compare the performance of supervised and unsupervised learning methods for place recognition \cite{Lowry2016_supervised}.
\citet{Krajnik2017_image} evaluated robustness of image feature to seasonal appearance variations over five datasets covering oceanic and continental climates.
}
In contrast to visual methods, cross-season place recognition with ranging sensors has received less attention.
For instance, \citet{Cao2021} proposed converting 3D~point cloud data into 2D~cylindrical projections, applying Gabor filters before matching images with a nearest neighbor search.
The authors evaluated their method on the multi-season Oxford Robotcar dataset, together with a new dataset recorded in a monsoon-influenced humid continental (Dwa) climate zone.
More recently, the Wild-Places dataset~\cite{Knights2023} covers over one year of lidar and camera data in a humid subtropical (Cfa) climate zone.
Although utilized by several recent studies~\cite{Ramezani2023, Griffiths2025}, the challenges in Wild-Places arise primarily from unstructured forest geometry rather than seasonal shifts, as the humid subtropical climate experiences relatively minimal seasonal transformation.
While prior works are often limited by mild seasonal transformations or single-sensor modalities, we include a report on the performance of visual, lidar, and radar place recognition across the full year in the subarctic (Dfc) climate zone.

\section{METHODOLOGY}

\begin{figure}[htbp]
    \centering
    \includegraphics[width=\linewidth]{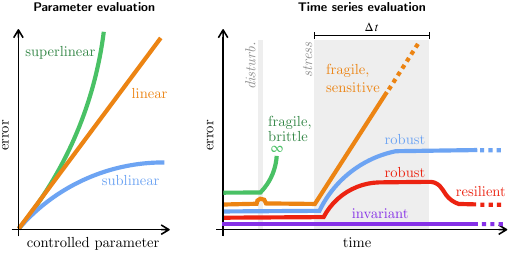}
    \caption{Illustration of the main differences between evaluation focusing on parameters as opposed to systems as a whole.
    \emph{Left}: Parameter evaluations highlight behaviors in controlled environments.
    \emph{Right}: Evaluation targeting system behaviors in an uncontrolled environment where disturbances (disturb.) and stresses will influence the localization errors, which are harder to control.}
    \label{fig:methodology:viability}
\end{figure}

As robots transition from indoor and urban environments into the unstructured outdoors, the cost of localization failure rises due to increased maintenance and operational overhead, particularly in remote locations.
Therefore, it is important to explicitly distinguish between algorithm evaluations, which are more common in the robotic literature, and field evaluations, the scope of this paper.
As shown in \autoref{fig:methodology:viability}-\emph{Left}, a typical laboratory setup evaluates algorithms on a set of controlled parameters while keeping the others constant.
The outcomes of such evaluations will describe the relations between the controlled parameters and an error metric, typically a metric related to path tracking error.
When a relation is identified, the algorithm can be described, in order of quality, as sublinear, linear, or superlinear.
Moreover, most of the evaluations will propose an overall comparison between the errors of state-of-the-art algorithms, which is often the average over all experiments on a standard dataset.
We will refer to these reported values as the \emph{nominal} performances and assume that the original authors were diligent in demonstrating \emph{reproducibility} (\ie~the repeated use of their code, data, and evaluation methods led to the same results).

Therefore, the focus of this paper is on \emph{replicability} (\ie~confirmation from independent studies) through field experiments.
Analyzing results from field experiments brings specific challenges and uncertainties, leading to a shift in methodology towards system-wide evaluations.
In this section, we first bring clarity to evaluation protocols in the domain of field robotics with the intention to propose a framework for reporting results.
Second, we present a methodology for evaluating localization and mapping in multi-season unstructured environments.
Finally, we explain our evaluation metrics related to field experiments, as they can bring their own challenges when compared to controlled environments.

\subsection{TAXONOMY OF TIME SERIES EVALUATION}
\label{section:methodology:time_series_evaluation}

At a high level, we observe a given system with respect to an environment.
In a long-term deployment, both the robot and the environment will change through time, generating \emph{perturbations} potentially leading to performance degradations.
Perturbations can be divided into two types lying on a continuous spectrum: 1) \emph{disturbances}, which are short but meaningful events; 2) \emph{stressors}, which affect the system over a long period of time.
In \autoref{fig:methodology:viability}-\emph{Right}, this difference is represented with the stress having a larger duration $\Delta t$.
Moreover, we give a list of examples and their potential causes encountered in our field deployments in \autoref{tab:stressors} to illustrate the difference between these two types.

\begin{table}[bht]
\centering
\caption{Examples of perturbations (\ie~stressors and disturbances) observed in our field deployments.}
\label{tab:stressors}
\begin{tabularx}{\columnwidth}{@{}llX@{}}
\toprule
\textbf{Perturbation} & \textbf{Type} & \textbf{Potential cause} \\ \midrule
Collisions & D & Driver inattention; vegetation blocking path \\
Hardware failure & D & Cold temperature; wear and tear; accidents \\
Software failure & D & Integer overflow; memory leaks \\
Rain/precipitation & S/D &  Humid climate zones \\
Wheel slippage & S/D & Ice, snow; dirt; wet and soft terrain \\
Env. changes & S & Location of the task or mission \\
Hardware changes & S & Changing vehicle or components \\
Signal changes & S & Replacing sensors \\
\rev{Luminosity changes} & S\rev{/D} & Daily and seasonal changes; \rev{occlusions} \\
Ground roughness & S & Snow accumulation; ruts \\
Real-time & S & Vehicle velocity; timely decision \\
\bottomrule
\multicolumn{3}{l}{\footnotesize{\emph{Legend}: D = disturbance, S = stressor, Env. = Environment}}
\end{tabularx}
\end{table}

Having a system subject to random perturbation, we want to characterize the expected relation between the two.
From here, we will use the term performance as the inverse of the error, and use both terms depending on the situation.
As highlighted in \autoref{fig:methodology:viability}, we can transfer the vocabulary from continuum mechanics and characterize this relation by giving an explicit definition of fragility, robustness, and resilience.
For these three relations, the analysis focuses on variation over the nominal performances and their trends in time.
Firstly, a \emph{fragile} system will fail shortly after a disturbance or under stress.
These failures are often explained by a perturbation bringing the system outside its working assumptions, whether they were explicitly stated by the original authors or not.
When the error response to perturbation is superlinear, the costs of failing fragile systems become high in a deployment.
For some fragile systems, it is possible to observe a rapid increase in error over time, while others will fail without warnings.
The latter is considered a \emph{brittle} system (\ie~rigid but fragile) and is the most risky to deploy.
Within the family of fragile systems, we define \emph{sensitive} systems as exhibiting a linear growth of error when under stress.
These systems degrade continuously, thus not ensuring the viability of the deployment.
Secondly, a \emph{robust} system is used to describe a capability to maintain in time a degree of performance under perturbation.
An \emph{invariant} system to a given perturbation would be an extreme case of robustness.
In our case, we define maintaining performance as an error growth over time that is sublinear.
\rev{Robustness should be evaluated over a sufficiently long duration to allow system performance to stabilize.
Shorter assessments risk misclassifying a fragile system as stable.}
Finally, \emph{resilient} systems can recover after it encounters a perturbation.
This recovery can be characterized on a spectrum between plastic and elastic.
A fully \emph{plastic} system would not be able to recover any error after the perturbation, as a fully \emph{elastic} system would bring the error back to its nominal value.
Similarly to a robustness evaluation, resilience can be identified only over a long period of recovery.

\rev{
It is important to note that bridging these theoretical definitions with empirical metrics is not straightforward.
As off-road environments lack the repeatability of controlled laboratory settings, stressors and disturbances often compound, making it difficult to isolate variables and assign definitive, quantitative labels.
Therefore, the goal of introducing these terms is to establish a shared, qualitative vocabulary.
We aim to identify trends and behavioral patterns that perturbations have on different variants of odometry, localization and mapping algorithms, rather than definitively classifying an algorithm into a specific taxonomic category.
}

\subsection{TAXONOMY OF LOCALIZATION AND MAPPING METHODS ACROSS SEASONS}
\label{section:methodology:taxonomy_loc_and_map}
In robotics literature, a typical evaluation of localization and mapping algorithms includes an analysis of the effects of controlled parameters on the algorithm's performance, usually followed by a comparison with state-of-the-art methods of the same category (\eg~sensor type or compute costs).
In this work, we argue that a field evaluation should take a wider perspective, as the gains of a method can be over-weighted by its increased complexity and, sometimes, fragility.
To this extent, we identify four categories of localization and mapping methods pertinent to multi-season outdoor environments.
This taxonomy is motivated by an example of an autonomous \ac{UGV} performing a resupplying mission, repeating the same trajectories throughout the year.
The four categories are expected to have the following behavior:

\begin{itemize}
    \item \emph{Proprioceptive odometry}: Relying only on wheel encoders and \acp{IMU}, proprioceptive systems are invariant to appearance changes of the environment as long as the system's motion model is well tuned and the sensors' biases are known.
    Therefore, we expect proprioceptive odometry to be more robust against long-term stressors (e.g., snow covering the ground) than short disturbances such as wheel slippage.
    Undoubtedly, relying solely on proprioception is insufficient for many navigation tasks (e.g., obstacle detection, path planning).
    Nevertheless, when focusing on pose estimation, recent evidence suggests that proprioceptive algorithms can perform on par with state-of-the-art exteroceptive methods at a fraction of the computation costs~\cite{LeGentil2025, Diener2025}.

    \item \emph{Exteroceptive odometry}: Adding external measurements from cameras, lidar, or, more recently, radar, sensors aim at correcting the unavoidable drift of the pure proprioceptive method.
    However, this correction comes at the cost of increased risk of fragility due to noisy sensor readings or an adverse environment.`
    Well-known examples where exteroceptive sensors struggle include precipitation~\cite{Courcelle2023_importance} or degenerate environments such as tunnels~\cite{Tuna2024_xICP}.

    \item \emph{Exteroceptive odometry with \ac{PGO}}: Methods incorporating loop closure search followed by \ac{PGO} aim at further correcting the robot's poses and improving the quality of the map by performing error back-propagation over the state space.
    However, these systems often rely on hand-tuned thresholds for loop closure detection, as well as estimation of localization uncertainty, which can be difficult to assess~\cite{Landry2019_cello}.
    In some cases, loop closure detectors add unnecessary complexity without any feasible gain in performance (\ie~when the performed trajectory does not contain any loops).

    \item \emph{Localization along a path}: Finally, when performing a localization task, it is possible to rely on a prior map or other representation of the environment.
    Then, the goal of localization becomes to track the robot's pose inside the environment with current sensor readings.
    Knowing the starting pose of the robot inside the environment simplifies the task, eliminating the need to solve the \emph{kidnapped robot problem}.
    As the costs of creating the environment representation can be high in terms of computing and expert time, it might be more advantageous to reuse an existing map instead of executing a full \ac{SLAM} pipeline for each autonomous deployment.
    However, any disturbances or stressors that appeared in the original map necessarily influence the subsequent localization task.
    \rev{It is worth noting that path-following localization is also achievable with purely appearance-based systems \cite{Krajnik2010_simple, vanDijk2024_visual}, although their repeat-phase performance is highly sensitive to appearance consistency across traversals.}
\end{itemize}

\begin{figure}[htb]
    \centering
    \includegraphics[width=0.8\linewidth]{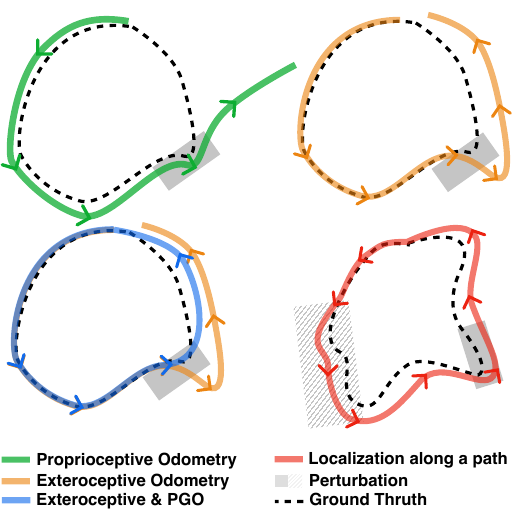}
    \caption{Example taxonomy of localization and mapping algorithms in multi-season outdoor environments.
    The figure depicts the expected behavior of each method, where the method's deviation from the \acl{GT} decreases with raising complexity.
    Green, orange, and blue lines originate from the same data, while the red line represents localization at a later date.}
    \label{fig:methodology:local_and_mapping}
\end{figure}

To summarize this conventional view on localization and mapping, we expect each level of complexity to constrain the effects of perturbations on the performance, as illustrated in \autoref{fig:methodology:local_and_mapping}.
Proprioceptive odometry (green line) is expected to be fragile to a perturbation (gray areas), whereas exteroceptive odometry (orange line) is expected to lower the drift.
Loop closure and \ac{PGO} performs global optimization over the whole trajectory (blue line), propagating new constraints into the graph structure and correcting drift along the way.
This type of solution is expected to lower the path-tracking error when compared to the exteroceptive odometry trajectory (orange line).
Last but not least, localization in prior maps anchors the robot's pose in the existing environment representation.
Nevertheless, scene changes degrade performance, as the sensor measurements lack valid correspondences in the map.
This change is denoted in the lower right corner of \autoref{fig:methodology:local_and_mapping} by the dashed area.
Given that structural alterations (\eg~seasons, wildfires, human activities) are common in subarctic environments, it could be advantageous to employ full exteroceptive odometry with \ac{PGO} for every deployment.
However, making a system more robust by adding complexity might be undesirable, as the extra performance can be coupled with higher fragility.
\rev{Additionally, complex systems come with additional development, testing and maintenance costs.}
Although certain missions, such as search-and-rescue operations, undoubtedly require full simultaneous localization and mapping pipelines, most current activities in the subarctic regions are of a different nature.
Forestry and mining, the largest industries in the area, mainly operate on predefined routes.
The forestry industry is highly competitive, with an incentive to bring down hardware costs in environments where compute-intensive graph optimization needs to be performed on the edge, as cloud connectivity is limited.
Furthermore, a prior map has secondary advantages, such as the ability to estimate the amount of snow, which is crucial for mining operations with the spring thaw.

Applying this taxonomy, this report evaluates the performance of odometry, localization, and mapping solutions while analyzing the correlation between algorithmic complexity and system fragility.
For this evaluation, we need a set of evaluation metrics capable of capturing the errors stemming from the disturbances and stresses occurring in field deployments.
The next section presents an overview of such metrics in the context of autonomous robots deployed in remote locations.

\subsection{EVALUATION METRICS}

\label{section:methodology:evaluation_metrics}

In robotics, odometry and localization are typically evaluated against known \acf{GT} trajectories.
These reference trajectories contain the robot's positions and possibly orientations derived from external measurements, such as a~tracking \ac{RTS}, \ac{GNSS} positioning, or a \ac{MOCAP} system.
\ac{MOCAP} systems provide high-quality millimeter-precision tracking, but their usage is limited to indoor spaces.
\ac{RTS} also offer millimeter-precision tracking, however, a single \ac{RTS} provides only point measurements.
Although it is possible to obtain a full 6-\ac{DoF} pose using a triple-\ac{RTS} setup~\citep{Vaidis2023_global}, \acp{RTS} remain impractical for large-scale field deployments due to their limited line-of-sight.
Consequently, researchers often rely on \ac{INS}, combining data from \ac{GNSS}, gyroscopes, and accelerometers.
\rev{The reliance of \ac{INS} on accelerometers can break the independence assumption of \ac{GT} if the same accelerometer data are used in a localization pipeline \cite{Ceriani2009_rawseeds}.}
To this end, our \ac{GT} is reconstructed from three~\ac{GNSS} antennas and a static reference antenna, recording the corrections that are used for \ac{PPK}.
We merge the three trajectories into one using point-to-Gaussian matching against the known spatial geometry of the \ac{GNSS} receivers on the robot.
Consequently, this process yields a reference trajectory containing only positions $\mathbf{q}_1\dots\mathbf{q}_n\in\mathbb{R}^3$, where $n$ is the cardinality.
Lacking explicit orientations, this \ac{GT} trajectory serves as the foundation for evaluating the translational components of the estimated 6-\ac{DoF} poses.

Our goal is to quantify the quality of the estimated trajectory $\mathbf{P}_1\dots\mathbf{P}_n\in\text{SE(3)}$ with respect to the reference \ac{GT} trajectory.
For brevity, we assume that the two trajectories are time-synchronized and of equal length.
To measure the estimate's quality, the community commonly relies on \ac{ATE} to assess global consistency and \ac{RTE} to measure local drift~\cite{Sturm2012_benchmark}.
Standard \ac{ATE} measures the absolute distance between trajectories, requiring an initial alignment $\mathbf{S}\in\text{SE(3)}$ typically found via least-squares optimization~\cite{Horn1987closed}.
When the \ac{GT} provides full 6-\ac{DoF} poses $\mathbf{Q}_i$, the error at index $i$ is defined as $\mathbf{E}_i=\mathbf{Q}^{-1}_i\mathbf{S}\mathbf{P}_i$.
The total \ac{ATE} is then computed as the mean error over the translational components:
\begin{equation}
    \text{ATE} = \frac{1}{n}\sum^n_{i=1}\left\|\text{trans}(\mathbf{E}_i)\right\|.
\end{equation}
An inherent characteristic of \ac{ATE} is that it exaggerates rotational errors happening at the beginning of the trajectory.
Additionally, it relies heavily on the initial alignment $\mathbf{S}$.
This alignment is often obtained by matching trajectory centroids, a process that can compromise the evaluation if the trajectories are expected to share an identical initial position and then drift apart.
Not relying on an initial global alignment $\mathbf{S}$, \ac{RTE} measures the local drift over a sliding window of $\Delta$ indexes.
At index $i$, the error $F_i$ is defined as:
\begin{equation*}
    \mathbf{F}_i = (\mathbf{Q}^{-1}_i\mathbf{Q}_{i+\Delta})^{-1}(\mathbf{P}^{-1}_i\mathbf{P}_{i+\Delta}).
\end{equation*}
While $\mathbf{F}_i$ is typically computed for multiple sizes of the sliding window, we maintain a general $\Delta$ for simplicity.
To normalize this error by the spatial scale of the window, we define the straight-line distance between the \ac{GT} endpoints as $d(i, i+\Delta) = \|\mathbf{q}_{i+\Delta} - \mathbf{q}_i\|$.
The total \ac{RTE} is derived by computing statistics, such as the mean, across all windows normalized by $d(i, i+\Delta)$, as
\begin{equation}
    \text{RTE} = \frac{1}{n}\sum_{i=1}^n \left\{ \frac{\left\| \text{trans}(\mathbf{F}_i) \right\|}{d(i, i+\Delta)} \right\}.
\end{equation}
The alignment for \ac{RTE} occurs independently per window, which strictly requires full 6-\ac{DoF}~poses for both the estimate and the \ac{GT} trajectories.
A common workaround when orientation is unavailable is to use metrics that avoid window alignment entirely, such as the ratio of estimated and true traveled distances.
Defining $s(i,~{i+\Delta})$ as the path length between index $i$ and $i+\Delta$, the \ac{RTDE} is computed as the mean ratio between the estimate and the \ac{GT} path lengths as
\begin{equation}
    \text{RTDE} = \frac{1}{n}\sum_{i=1}^n \left\{ \frac{s_e(i, i+\Delta)}{s_{gt}(i, i+\Delta)} \right\}.
\end{equation}
While useful for scale evaluation, this metric ignores rotational errors and can disguise fragile systems as robust.
For example, an estimator erroneously reporting a constant forward velocity while the robot is turning would result in a near-perfect \ac{RTDE}.
Furthermore, computing the relative error on consecutive and non-overlapping windows might ignore events happening near the trajectory end due to the boundary effect (\ie~remaining distance shorter than the required window length).

To address the absence of orientations in our \ac{GT} data, we define the \ac{SARTE} as a replacement for \ac{RTE}.
Let $\Delta_j$ denote the $j$-th sliding window of length $\Delta$ indices, where $j \in \{1, \dots, K\}$ and $K$ is the total number of extracted windows.
For each window, we find a local rigid-body transformation $\mathbf{S}_j=\left(\mathbf{R}_j,\mathbf{t}_j\right)$.
For the initial window ($j=1$), the estimate $\mathbf{P}$ and the \ac{GT} $\mathbf{Q}$ rely on an initial global alignment $\mathbf{S}_1=\mathbf{S}$.
This global alignment is obtained by fixing the initial position of the two trajectories and finding the optimal rotation to minimize their distance using the Kabsch algorithm.
For all subsequent windows ($j>1$), we isolate the final \qty{50}{\percent} of the positions from the preceding window $\Delta_{j-1}$ to compute a local rigid-body transformation $\mathbf{S}_j$.
This transformation aligns the estimate trajectory segment to the \ac{GT}, effectively compensating for the lack of orientation in the \ac{GT} data.
The aligned estimated positions are given as $\mathbf{\hat{p}}_k=\mathbf{S}_j\left(\mathbf{p}_k\right)$ for $k\in\left\{i\dots i+\Delta\right\}$.
We then compute the translational error against the corresponding \ac{GT} segment as the Euclidean distance between the two end points of the aligned estimate and reference trajectories.
Omitting the index $j$ for brevity, the total \ac{SARTE} is reported as the mean of these Euclidean distances, normalized by the distance $d(i, i+\Delta)$, across all windows:
\begin{figure}[t]
    \centering
    \includegraphics[width=0.98\linewidth]{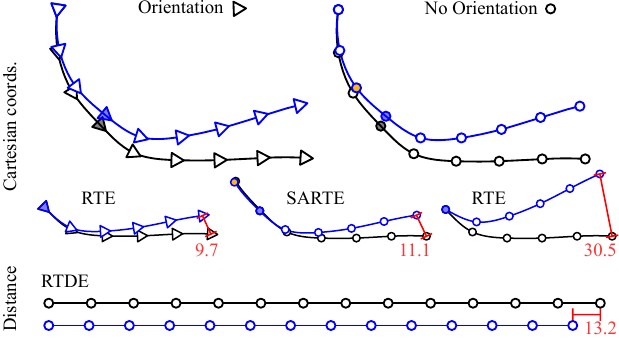}
    \caption{\rev{An illustration of the discussed error metrics.
    \acf{RTE} requires orientations to correctly align the two sub-trajectories.
    \acf{SARTE} uses part of the previous window as a proxy to estimate the orientation.
    \acf{RTDE} is distance-based and does not account for errors in orientation.
    Numerical values in this toy example correspond the red lines lengths.}}
    \label{fig:methodology:error_metrics}
\end{figure}
\begin{equation}
    \text{SARTE} = \frac{1}{n}\sum_{i=1}^n \left\{ \frac{\left\| \mathbf{q}_{i+\Delta} - \mathbf{\hat{p}}_{i+\Delta} \right\|}{d(i, i+\Delta)} \right\}.
\end{equation}
While \ac{SARTE} provides a robust workaround for position-only reference data, it has inherent limitations.
The metric is constrained by the \ac{GT} quality and the trajectory shape relative to the minimal window size $\Delta$.
If $\Delta$ is too small, locally non-smooth \ac{GT} segments or ambiguities in rotation around the trajectory axis can result in poorly constrained alignments and, consequently, an inaccurate evaluation.
\rev{We illustrate the differences between \ac{RTDE}, \ac{RTE} and \rev{SARTE} in \autoref{fig:methodology:error_metrics}.}

Reporting a relevant error metric becomes increasingly complex in the case of \emph{fragile} methods.
\rev{Therefore, we also document the number of failures for each method-trajectory pair, and we report the \emph{failure rate}, \ie~the ratio of the failed and total number of runs.}
An estimated trajectory is classified as a failure, and the method that produced it is called \emph{fragile}, in case of an premature termination of odometry or localization algorithm.
\rev{
As our trajectories are time synchronized, we detect this early termination by comparing the last estimate and \ac{GT} timestamps $t_\text{est}$ and $t_\text{GT}$.
A run is labeled as failure if $t_\text{est} < 0.95~t_\text{GT}$.
}
In contrast, while severe state estimation degradation can compromise an autonomous mission, we do not count it as a failure as it is reflected in the other error metrics.
Finally, we introduce the \ac{LFR}~[\qty{}{\percent}] to quantify performance during localization in a prior map.
\rev{Defining $d_e$ as the total trajectory duration and $d_f$ as the time until localization failure (using the above failure definition), the \ac{LFR} is the complementary ratio of successful localization relative to the total sequence duration:
\begin{equation}
    \text{LFR} =
    \begin{cases}
        1 - \dfrac{d_f}{d_e} & \text{if failure occurs} \\[1.5ex]
        0 & \text{otherwise.}
    \end{cases}
\end{equation}
}
The metric therefore quantifies the portion of the trajectory lost due to localization tracking failure.
For example, \qty{0}{\percent} corresponds to a failure-free run, while \qty{100}{\percent} indicates an immediate loss of localization.

\begin{figure*}[!t]
    \centering
    \includegraphics[width=0.99\linewidth]{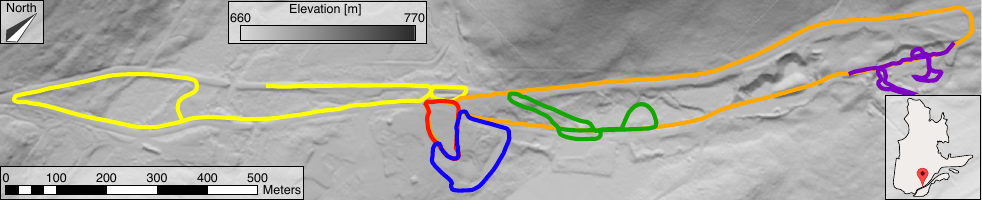}
    \caption{\rev{Elevation map of the deployment site with the six evaluated trajectories.
    The bottom-right inset shows the location of the site within the province of Quebec, Canada.}
    }
    \label{fig:trajectories}
\end{figure*}

\section{EXPERIMENTAL SETUP}
\label{section:experiments}

In this work, we report the findings of our 12-month field campaign, covering more than \qty{64}{\kilo\meter} of sensor data recorded under on-road, off-pavement, and off-trail conditions in the subarctic (Dfc) climate zone.
\rev{
We start by briefly describing the field campaign and data acquisition process, before introducing the nine evaluated methods.
While this work details a thorough evaluation of the odometry, localization and mapping approaches, as well as the challenges and lessons learned during the field campaign, the data collection itself was exhaustively detailed in a prior publication~\cite{Boxan2026_fomo}.
}

\subsection{DATA RECORDING}
\rev{
Our deployments were conducted with a half-tonne Clearpath Warthog \ac{UGV}, equipped with a front-facing ZED~X stereo camera, a rear-facing Basler monocular camera, a 128-channel RoboSense Ruby Plus lidar, a Navtech CIR-304H \ac{FMCW} radar, three Emlid Reach~M2 \ac{GNSS} receivers, and a VectorNav VN100 \ac{IMU}.
The \ac{UGV} features a \qty{7.8}{\kilo\watt\hour} lithium-ion battery, and was mounted on tracks between January and March to improve mobility on snow.
Manual control was maintained for most of the runs, while minority of sequences were executed via a \acl{LTR} framework using waypoints from a prior deployment.
The \acl{GT} data originate from the three \ac{GNSS} receivers mounted on the \ac{UGV} and are post-processed with a maintained \ac{PPK} base station.
The six repeated sequences, or trajectories, mainly take place on wide forest roads with good sky visibility.
More details on the \ac{GT} generation can we found in~\cite{Boxan2026_fomo}.
We refer to the six trajectories, depicted in \autoref{fig:trajectories}, with a color code as \texttt{Red}, \texttt{Blue}, \texttt{Green}, \texttt{Yellow}, \texttt{Orange}, and \texttt{Magenta}.
The sequences have the following particularities:
\begin{itemize}
    \item \texttt{Red} is the shortest covering \qty{300}{\meter} meters of on-road, semi-urban loop;
    \item \texttt{Blue} spans more than \qty{500}{\meter} of mixed on-road and off-road sections, leading the robot downhill into a forest and back to its starting point through heavy snow in winter;
    \item \texttt{Green} consists of over \qty{500}{\meter} of off-trail navigation characterized by steep elevation changes and dense vegetation that limits camera visibility;
    \item \texttt{Orange} is the longest at \qty{2200}{\meter}, combining forest roads and a stone quarry section;
    \item \texttt{Yellow} covers \qty{1900}{\meter} of on-road and off-road navigation and contains two loop closures; and finally
    \item \texttt{Magenta} provides an approximately \qty{700}{\meter} off-trail loop through the stone quarry, featuring irregular terrain and common collisions with trees and boulders.
\end{itemize}
While the \texttt{Red} trajectory was repeated in twelve data recordings, \texttt{Blue}, \texttt{Green}, and \texttt{Magenta} were featured ten times, and the longest trajectories (\ie~\texttt{Yellow} and \texttt{Orange}), nine times.
All trajectories contain at least three deployments in winter, summer and autumn.
}
\subsection{EVALUATED METHODS}



\newcommand{\refsProprio}{
    \shortautoref{section:odometry-in-subarctic-environments}{\ref*{section:results-subarctic-performance-report}},
    \shortautoref{section:real-time-evaluation}{\ref*{section:results-subarctic-performance-report}}
}
\newcommand{\refsWILN}{
    \shortautoref{section:odometry-in-subarctic-environments}{\ref*{section:results-subarctic-performance-report}},
    \shortautoref{section:real-time-evaluation}{\ref*{section:results-subarctic-performance-report}},
    \shortautoref{section:seasons-localization}{}
}
\newcommand{\refsLamma}{
    \shortautoref{section:odometry-in-subarctic-environments}{\ref*{section:results-subarctic-performance-report}},
    \shortautoref{section:results-loop-closure-subarctic}{\ref*{section:results-subarctic-performance-report}},
    \shortautoref{section:real-time-evaluation}{\ref*{section:results-subarctic-performance-report}},
    \shortautoref{section:seasons-localization}{}
}
\newcommand{\refsORB}{
    \shortautoref{section:odometry-in-subarctic-environments}{\ref*{section:results-subarctic-performance-report}},
    \shortautoref{section:results-loop-closure-subarctic}{\ref*{section:results-subarctic-performance-report}},
    \shortautoref{section:real-time-evaluation}{\ref*{section:results-subarctic-performance-report}},
    \shortautoref{section:results-seasons-sensing}{},
    \shortautoref{section:visual-feature-detection}{\ref*{section:results-seasons-sensing}},
    \shortautoref{section:seasons-localization}{}
}
\newcommand{\refsPyCu}{
    \shortautoref{section:odometry-in-subarctic-environments}{\ref*{section:results-subarctic-performance-report}}
}
\newcommand{\refsDROID}{
    \shortautoref{section:odometry-in-subarctic-environments}{\ref*{section:results-subarctic-performance-report}},
    \shortautoref{section:results-loop-closure-subarctic}{\ref*{section:results-subarctic-performance-report}},
    \shortautoref{section:real-time-evaluation}{\ref*{section:results-subarctic-performance-report}},
    \shortautoref{section:visual-feature-detection}{\ref*{section:results-seasons-sensing}},
}
\newcommand{\refsKISS}{
    \shortautoref{section:odometry-in-subarctic-environments}{\ref*{section:results-subarctic-performance-report}},
    \shortautoref{section:results-loop-closure-subarctic}{\ref*{section:results-subarctic-performance-report}},
    \shortautoref{section:real-time-evaluation}{\ref*{section:results-subarctic-performance-report}},
    \shortautoref{section:results-seasons-sensing}{},
    \shortautoref{section:seasons-localization}{}
}
\newcommand{\refsRTR}{
    \shortautoref{section:odometry-in-subarctic-environments}{\ref*{section:results-subarctic-performance-report}},
    \shortautoref{section:results-loop-closure-subarctic}{\ref*{section:results-subarctic-performance-report}},
    \shortautoref{section:real-time-evaluation}{\ref*{section:results-subarctic-performance-report}},
    \shortautoref{section:results-seasons-sensing}{},
    \shortautoref{section:seasons-localization}{}
}
\newcommand{\refsNavtech}{
    \shortautoref{section:odometry-in-subarctic-environments}{\ref*{section:results-subarctic-performance-report}},
    \shortautoref{section:results-loop-closure-subarctic}{\ref*{section:results-subarctic-performance-report}},
    \shortautoref{section:real-time-evaluation}{\ref*{section:results-subarctic-performance-report}},
}

\begin{table*}[ht]
\centering
\caption{Summary of the evaluated methods.}
\label{tab:methods}
\begin{tabularx}{\textwidth}{@{}lXlcccccccc@{}}
\toprule
\textbf{Year} & \textbf{Method} & \textbf{Ref.}   & \textbf{Wheel encoders} & \textbf{\ac{IMU}}  & \textbf{Stereo cam.} & \textbf{Lidar} & \textbf{Radar} & \textbf{LC} & \textbf{Loc. mode} & \textbf{Dim} \\ \midrule
2011 & Proprioceptive &~\cite{Madgwick2011}             & \used     & \used    & \notused & \notused & \notused & \notused & \notused & 3D\\
2022 & WILN &~\cite{Baril2022}           & \used     & \used    & \notused & \used    & \notused & \notused & \used    & 3D\\
2025 & 2Fast-2Lamaa         &~\cite{LeGentil2025_lamaa}  & \notused  & \used    & \notused & \used    & \notused & \used    & \used    & 3D \\
2021 & ORB-SLAM3 &~\cite{Campos2021_orbslam3}                     & \notused  & \used    & \used    & \notused & \notused & \used    & \used    & 3D \\
2025 & cuVSLAM &~\cite{Korovko2025cuvslam}            & \notused  & \notused & \used    & \notused & \notused & \used    & \used    & 3D \\
2021 & DROID-SLAM &~\cite{Teed2021}                     & \notused  & \notused & \used    & \notused & \notused & \used    & \notused & 3D \\
2025 & KISS-SLAM &~\cite{Guadagnino2025}                & \notused  & \notused & \notused & \used    & \notused & \used    & \notused    & 3D \\
2025 & RT\&R &~\cite{Qiao2025}                       & \notused  & (Gyro)   & \notused & \notused & \used    & \notused & \used    & 2D \\
2023 & Navtech-Radar-SLAM &~\cite{Lim2023}              & \notused  & \notused & \notused & \notused & \used    & \used    & \notused & 2D  \\
\bottomrule
\multicolumn{8}{l}{\footnotesize{\emph{Legend}: Cam. = Camera, LC = Loop closure, Loc. mode = Possible to run the algorithm to localize along a prior path.}}
\end{tabularx}
\end{table*}

We focus on the effect of perturbations on nine odometry and localization algorithms, summarized in \autoref{tab:methods}.
\rev{
Among these nine methods, each sensor modality,~\ie~lidar, radar, and camera, is represented at least two times.
While we only considered recent and community-tested methods, we attempted to increase variability by incorporating methods with different underlying operating principles,~\eg~classic and deep learning-based approach for feature extraction for visual odometry methods.
The nine selected methods are also all open-source.
}
Firstly, we have the proprioceptive approach, integrating wheel angular velocities to estimate the forward displacement and a Madgwick filter~\cite{Madgwick2011} for orientation estimation from \ac{IMU} data.
Employing the RoboSense lidar, WILN~\cite{Baril2022} is an \ac{ICP}-based localization and mapping algorithm using data from the proprioceptive method as its prior.
\rev{WILN was previously tested in similar conditions of a boreal forest.}
\rev{While KISS-SLAM~\cite{Guadagnino2025} also relies on point-to-point \ac{ICP}, it performs \ac{PGO} to correct local maps, provided by a lidar-odometry algorithm with a constant velocity motion model~\cite{vizzo2023}.}
Also using the lidar sensor, 2Fast-2Lamaa~\cite{LeGentil2025_lamaa} performs localization and mapping using Gaussian distance field, with continuous \ac{IMU} pre-integration for lidar scan deskewing.
We configure the package to use local topographic maps, and we perform loop closure detection and \ac{PGO} in post-processing.
Leveraging the information from radar, \acl{RTR}~\cite{Qiao2025} also relies on topographic local maps, while only utilizing the 2D~\ac{FMCW} radar and a gyroscope.
The second evaluated radar method is Navtech-Radar-SLAM, based on the radar odometry ORORA~\cite{Lim2023}, and ScanContext for loop-closing~\cite{Kim2018}.
\rev{Unlike \ac{RTR}, ORORA decouples rotation and translation estimation.}
Finally, we also include three methods employing a stereo camera.
ORB-SLAM3~\cite{Campos2021_orbslam3} is a \ac{VSLAM} feature-based method, relying on Maximum-a-Posteriori estimation and incorporating readings from an \ac{IMU}.
Similarly, cuVSLAM~\cite{Korovko2025cuvslam} utilizes hardware-accelerated feature extraction and tightly-coupled optimization on the GPU.
DROID-SLAM~\cite{Teed2021}, on the other hand, is a differentiable approach refining the robot's path with a recurrent neural network.
Localization along a path while reusing a prior map is supported by five of the nine evaluated methods: WILN, 2Fast-2Lamaa, ORB-SLAM3, cuVSLAM, and \ac{RTR}.
Conversely, the proprioceptive approach inherently lacks a map reference, rendering absolute localization impossible.
DROID-SLAM is also excluded, as its architecture relies on a dense, dynamically optimized map rather than a static structure.
Furthermore, localization is not implemented in KISS-SLAM, primarily because its underlying KISS-ICP odometry operates on a sliding temporal submap rather than a persistent global representation
Similarly, Navtech-Radar-SLAM lacks a localization mode and constructs a global map, which is susceptible to radar drift and noise.
Additionally, all exteroceptive methods, except WILN and \ac{RTR}, detect loop closures for trajectory refinement.
We preserved the original parameters for all methods, performing a limited parameter search only if a method failed on the easiest trajectory, which is a \qty{300}{\meter} long loop with distinct urban structures.
Furthermore, DROID-SLAM parameters were specifically configured so that our longest data recording would fit within the 48~GB VRAM of the evaluation computer, equipped with an Nvidia Quadro~RTX~8000.

\section{RESULTS}
\label{section:results}

Having defined the field conditions, the evaluation metrics, and the localization methods, we focus our evaluation effort on three main questions:
\begin{enumerate}
    \item How do the various localization and mapping algorithms perform when subject to the stress induced by subarctic conditions, and environment well outside of the typical laboratory setup?
    \item How are the individual data modality influenced by the seasonal changes?
    \item How do the methods perform against a long-term stress, represented by seasonal changes, when localizing in an existing map?
\end{enumerate}

We will start with a nominal case, evaluating the pose accuracy on six diverse trajectories over twelve months in subarctic conditions.

\subsection{SUBARCTIC CONDITIONS AS A STRESSOR ON SLAM}
\label{section:results-subarctic-performance-report}

This section provides a performance overview of the evaluated methods when subjected to two primary stressors: sensor variations and changes in environmental conditions.
Both of these stressors will impact our reproducibility results when compared to the evaluation provided by the original authors.
Starting with the sensor change, feature extraction or pose tracking can be sensitive to many factors, including data rates, camera resolution, number of lidar points, or radar signal strength.
The challenges that subarctic areas bring include the lack of distinct features for place recognition or geometrically-degenerate environments such as long forest corridors.
In winter, high snow banks create a tunnel-like environment providing little constraints in the robot's forward direction.
Contrast between the high albedo of snow and the darker tree canopy produces a non-uniform surface albedo throughout the scene.
Additionally, the evaluated methods face numerous disturbances, from collisions to vegetation blocking the field of view to wheel slip.
To isolate these challenges from systemic issues, all methods were evaluated offline, processing every sensor data frame sequentially to prevent data loss from network lag, buffer overflows, or insufficient computational resources.
Following the motivation from \autoref{section:methodology:taxonomy_loc_and_map}, the offline evaluation refers to the mapping run, in which the \ac{UGV} collects data while traversing the route, and the map is later created during postprocessing.

\begin{figure*}[htbp]
    \centering
    \includegraphics[width=\linewidth]{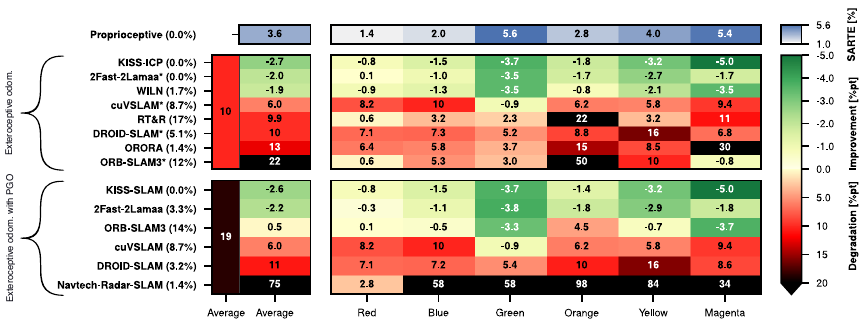}
    \caption{
        Reported mean \acf{SARTE} and improvement or degradation over Proprioceptive as percentage points (\qty{}{\pp}) for eleven evaluated methods over six trajectories in a subarctic environment.
        The failure rate of each method follows the method's name using (X \%).
        The first column shows the total degradation of a method type (Exteroceptive odometry, Exteroceptive odometry with \acl{PGO}) over Proprioceptive.
        The second column shows the total degradation of each method over Proprioceptive.
        The remaining values reported for each trajectory are averaged over all 12 recordings throughout the year.
        A star (*) denotes the odometry output of a method.
        }
    \label{fig:results-subarctic-performance}
\end{figure*}

The complete performance report is presented in \autoref{fig:results-subarctic-performance}.
The figure details the results for the nine evaluated methods across six trajectories, where each row represents the \ac{SARTE} for a specific method. 
Recall that we use \ac{SARTE} as a proxy for the standard \ac{RTE}.
We choose the size of the sliding windows $\Delta$ so that the straight line distance $d(i, i+\Delta)\in\{ 100, 150, \dots, 300 \}\,\unit{\meter}$.
Compared to the KITTI benchmark~\cite{Geiger2012_kitti}, our maximum window size is reduced to reflect the generally shorter length of our trajectories, while the lower bound is determined by the rotation ambiguity around the forward movement axis.
Following our methodology, the evaluated methods are divided into three categories: proprioceptive, exteroceptive, and exteroceptive with \acf{PGO}.
For methods featured in both exteroceptive categories, an asterisk (*) denotes the odometry-only version (\ie~without loop closure and \ac{PGO}).
From left to right, the figure's columns provide more details about the performance of each category and each method.
The first column presents an overall average \ac{SARTE} across all recordings and all trajectories for each category.
The second column shows the mean error for each individual method across all recordings and trajectories.
The next six columns detail the results for each specific trajectory, computed as the mean over all respective recordings.
The data presented in \autoref{fig:results-subarctic-performance} exclude complete failures, described in \autoref{section:methodology:evaluation_metrics}.
The failure rate, computed as the ratio of failures and the total number of runs, is presented after each method's name in parentheses.
\rev{We remind the reader that a failure is determined by the portion of the trajectory completed over time, rather than the magnitude of the reported error.
To provide supplementary context, partial qualitative results are presented in the \hyperref[section:appendix]{Appendix}.
Additionally, full results are available in a Docker-based visualization tool.\footnote{\url{https://anonymous.4open.science/r/subarctic_trajectory_visualizer}}
}

\subsubsection{ODOMETRY IN SUBARCTIC ENVIRONMENTS}
\label{section:odometry-in-subarctic-environments}

In \autoref{fig:results-subarctic-performance}, the values for the Proprioceptive odometry represent the \ac{SARTE}, whereas values for the other methods indicate their relative improvement or degradation compared to this baseline.
Improvements are reported as negative values and degradation as positive values, allowing readers to calculate the \ac{SARTE} for any method by adding its relative score to the Proprioceptive baseline row.
The Proprioceptive odometry establishes a surprisingly strong baseline, achieving a combined mean \ac{SARTE} of \qty{3.6}{\percent}.
This error is lowest on the short \texttt{Red} trajectory, which is fully paved and thus features low wheel slip.
Conversely, the highest \ac{SARTE} of \qty{5.6}{\percent} and \qty{5.4}{\percent} occur on the \texttt{Green} and \texttt{Magenta} trajectories.
These two routes contain the most off-trail segments with steep slippery hills, collisions and rough terrain.
Overall, the proprioceptive drift is a combination of gyroscope bias and wheel slip, occurring mostly in winter months on the uphill sections of \texttt{Green}, \texttt{Yellow}, and \texttt{Magenta} trajectories.
Compared to a similar study on Proprioceptive odometry by \citet{LeGentil2025} on a vehicle drifting sequence, their reported \rev{mean \ac{RTE} of \qty{1.0}{\percent} is almost} four times lower than our combined error of \qty{3.6}{\percent}.
We attribute the higher error to our Proprioceptive odometry operating in off-road~3D~environment, together with the fact that we deployed a skid-steered instead of an Ackermann-steering vehicle.

Moving to exteroceptive odometry, counter to our intuition, methods in this category typically performed worse than the Proprioceptive baseline, with the average degradation of \qty{10}{\pp}.
All lidar-based methods consistently improved upon the proprioceptive drift across most trajectories and showed similar improvements over the baseline.
While the improvement was the lowest on the short \texttt{Red} trajectory, the highest improvements were achieved on the off-trail \texttt{Green} and \texttt{Magenta} routes.
The only failure experienced by these methods is attributed to WILN on the \texttt{Green} trajectory, due to an integer overflow in the \ac{IMU} data.
Overall, only KISS-ICP lidar exteroceptive odometry method achieved a mean error lower than \qty{1}{\percent}.
These results are in contrast with literature values that typically fall below the \qty{1}{\percent} threshold (\eg~\qty{0.3}{\percent} for 2Fast-2Lamaa* on the Boreas dataset~\cite{Burnett2023_boreas}, \qty{0.61}{\percent} for KISS-ICP on KITTI odometry benchmark~\cite{Geiger2012_kitti}).
This performance degradation highlights the difficulty of lidar odometry in subarctic environments.
\acl{RTR} showed degraded performance on all trajectories, including the short paved \texttt{Red} route.
The degradation peaked at \qty{22}{\pp} on the long \texttt{Orange} route due to a drift in orientation, despite \ac{RTR} also utilizing gyroscope angular velocities.
The external source of angular velocities also did not lead to a good performance on the off-trail \texttt{Magenta} trajectory, where the \ac{UGV} experiences many quick on the spot turns.
Exhibiting a failure rate of \qty{17}{\percent}, \ac{RTR} proved to be the most fragile among all the evaluated exteroceptive odometry methods.
These failures mostly stem from a specific creek crossing on the \texttt{Orange} and \texttt{Magenta} routes.
As the robot goes downhill, a sudden change in pitch causes the employed \ac{CFAR} point extractor to register ground strikes on one side and zero returns on the other, where the radar faces the sky.
Ultimately, this caused the odometry to fail due to an insufficient number of points to match in the \ac{ICP} loop.
Unlike \ac{RTR}, ORORA operates without a gyroscope and consequently fails to accurately track orientation changes during sharp robot turns.
\rev{ORORA's combined mean \ac{SARTE} reaches \qty{13}{\percent}, far higher than the authors' results on the MulRan dataset~\cite{Kim2020_mulran}, where they reported a mean \ac{RTE} of \qty{4.05}{\percent}.}
\rev{
Moving to three visual-odometry methods, the only improvement over Proprioceptive was achieved by cuVSLAM* and ORB-SLAM3* on the \texttt{Green} and \texttt{Magenta} trajectories, respectively.
The last visual-odometry method, DROID-SLAM*, showed a degraded performance on all six trajectories, reporting a mean \ac{SARTE} over \qty{13}{\percent}.
The \qty{5.1}{\percent} of failure rate for DROID-SLAM* corresponds to a night deployment, where visual methods predictably struggled.
On average, cuVSLAM* performed the best among the tested visual-based methods, with an error rate of \qty{8.7}{\percent}.
Despite utilizing stereo camera imagery, visual systems clearly do not transfer well to subarctic environments, especially when contrasted with the \qty{0.6}{\percent} mean error reported for ORB-SLAM3* on the EuRoC dataset \cite{Burri2016_euroc}, and \qty{0.29}{\percent} reported by cuVSLAM authors on the same benchmark \cite{Korovko2025cuvslam}.
}

\subsubsection{LOOP CLOSURES IN SUBARCTIC ENVIRONMENTS}
\label{section:results-loop-closure-subarctic}
Loop closures improve global trajectory consistency by recognizing visited locations, but they introduce complexity and potential fragility from false positives.
In this section we analyze how the performance of the evaluated exteroceptive odometry methods evolved after loop closure detection and \acf{PGO}.

As shown in the bottom part of \autoref{fig:results-subarctic-performance}, integrating loop closing and \ac{PGO} improved the state estimation for two out of the six evaluated methods.
\rev{A comparison of the average improvement, together with the number of expected and detected loop closures, is presented in \autoref{tab:loop_closure_stats}.}
\begin{table}[tb]
\centering
\caption{Loop closure statistics across trajectories.
Reported improvement or degradation as percentage points (\qty{}{\pp}) of \acl{SARTE} over the odometry counterparts for \rev{six} evaluated methods.
The number in parentheses after the performance change indicates the number of detected loop closures.
\rev{The number of expected loop closures for each trajectory is presented in parentheses in the first column.
For Magenta, the dash differentiates between the expected loop closure count for range sensors and visual methods.
}
}
\label{tab:loop_closure_stats}
\setlength{\tabcolsep}{2pt} 
\renewcommand{\arraystretch}{1.3} 
\scriptsize
\begin{tabular}{@{} l rrrrrr @{}}
\toprule
\addlinespace[3ex]
         & \rot{2Fast-2Lamaa} & \rot{ORB-SLAM3} & \rot{\rev{cuVSLAM}} & \rot{KISS-SLAM} & \rot{DROID-SLAM} & \rot{NR-SLAM} \\
\midrule
Red \rev{(12)}     & \rev{-0.4} (0)   & \rev{-0.5} (9)  & 0.0 (0) & \rev{0.0} (0) & \rev{0.0} (242) & \rev{-3.6} (14) \\
Blue \rev{(9)}    & \rev{-0.1} (0)   & \rev{-5.8} (8)  & 0.0 (0) & 0.0 (0) & \rev{-0.1} (600) & \rev{52.2} (1.1k) \\
Green \rev{(21)}   & \rev{-0.3} (1)   & \rev{-6.3} (10) & 0.0 (0) & 0.0 (0) & \rev{0.2} (4.7k) & \rev{54.3} (1.5k)  \\
Orange \rev{(9)}  & \rev{-0.1} (0)   & \rev{-45.5} (3)  & 0.0 (1) & \rev{0.4} (7) & \rev{1.2} (22k) & \rev{83} (3.7k)  \\
Yellow \rev{(19)}  & \rev{-0.2} (3)   & \rev{-10.7} (9)  & 0.0 (6) & \rev{0.0} (3) & \rev{0.0} (22k) & \rev{75.5} (2.8k) \\
Magenta \rev{(0/7)} & \rev{-0.1} (18)   & \rev{-2.9} (3)  & 0.0 (5) & \rev{0.0} (0) & \rev{1.8} (3.3k) & \rev{4.0} (656)  \\
\midrule
\textbf{Average} & \textbf{\rev{-0.2}} & \textbf{\rev{-11.95}} & \textbf{0.0} & \textbf{\rev{0.4}} & \textbf{\rev{0.52}} & \textbf{\rev{44.2}}\\
\bottomrule
\addlinespace[1ex]
\multicolumn{7}{l}{\textit{Legend}: NR-SLAM = Navtech-Radar-SLAM}
\end{tabular}
\end{table}
2Fast-2Lamaa improved the mean performance of its odometry counterpart to \rev{\qty{-0.2}{\pp}}, bringing its \ac{SARTE} down to \rev{\qty{1.4}{\percent}}.
We attribute most of the improvements to the trajectory smoothing effect of \ac{PGO} rather than to the detected loop closures.
This is supported by the fact that the largest improvement occurs on the \texttt{Red} trajectory, where no loop closures were detected, whereas the smallest improvement occurs on \texttt{Magenta}, despite 18~detected loop closures.
We note that 2Fast-2Lamaa only found loop closures on trajectories that contain longer overlapping segments, namely \texttt{Green}, \texttt{Yellow} and \texttt{Magenta}.
Despite these improvements, the errors remain notably higher than the reported values of \qty{0.3}{\percent} for 2Fast-2Lamaa on the Boreas dataset.
The visual-based ORB-SLAM3 exhibited the most significant improvement among the evaluated methods employing \ac{PGO}, with its mean degradation reduced from \qty{22}{\pp} to \qty{0.5}{\pp}.
This improvement can be attributed to the 42~detected loop closures across the 60~runs.
\rev{cuVSLAM detected several loop closures in the \texttt{Orange}, \texttt{Yellow} and \texttt{Magenta} trajectories, but these detected junctions had no effect on the error values as the \ac{SARTE} remained unchanged.}
\rev{In contrast, KISS-SLAM actually showed a worse performance to its odometry counterpart KISS-ICP.
This degradation was primary caused by the \texttt{Orange} trajectory, where the method saw a \rev{\qty{0.4}{\pp}} degradation over its odometry counterpart.
Similarly, DROID-SLAM also performed worse than its version without \ac{PGO}, reporting degradation over Proprioceptive of \qty{11}{\pp}, compared to \qty{10}{\pp} for DROID-SLAM*.
}
While the majority of DROID-SLAM loop closures are false positives, they have limited impact on the reconstructed trajectory as the network assigns a confidence for each loop closure edge it detects.
Conversely, Navtech-Radar-SLAM demonstrated that adding a loop closing system can significantly decrease state estimation quality.
In this implementation, Navtech-Radar-SLAM utilizes ScanContext~\cite{Kim2018} as its place recognition engine to trigger loop closures.
This configuration improved the performance on the \texttt{Red} trajectory by \qty{3.6}{\pp} compared its odometry front-end, ORORA.
However, the performance degraded on the other 48~runs, where ScanContext identified a total of 9860~loop closures, the vast majority of which were false positives.
Because the loop closure confidence threshold in Navtech-Radar-SLAM is a fixed parameter, the pipeline lacked the robustness to filter these incorrect constraints, resulting in a \rev{\qty{44.2}{\pp}} performance drop compared to ORORA.
This degradation due to false positives can be attributed to the self-similarity of boreal forest roads, particularly from the perspective of a 2D~sensor with a coarse resolution.

\rev{Overall, these results indicate that significant work remains to reliably generalize the performance of loop closures to subarctic environments.}
Existing place recognition techniques are typically fine-tuned on urban environments with distinct features, but struggle with monotonous data or sensor occlusion that we commonly see in our sequences.
Additionally, place recognition performance remained consistently low across all seasons.
We attribute this to the self-similarity of the boreal forest, which appears to exceed the discriminatory capabilities of current techniques even in the absence of snow.
In this context, the difficulty of the environment effectively masks any additional seasonal variations.
Although global accuracy is vital for environmental monitoring, it brings additional complexity that is unnecessary for certain navigation tasks, such as \acl{TaR}.
In conclusion, our evaluation revealed that the subarctic environment, coupled with sensor transitions, critically challenges the robustness of established odometry and loop-closing frameworks, even when executed offline.

\subsection{SEASONS AS A STRESSOR ON SENSING MODALITIES}
\label{section:results-seasons-sensing}

\begin{figure}[bthp]
    \centering
    \includegraphics[width=\linewidth]{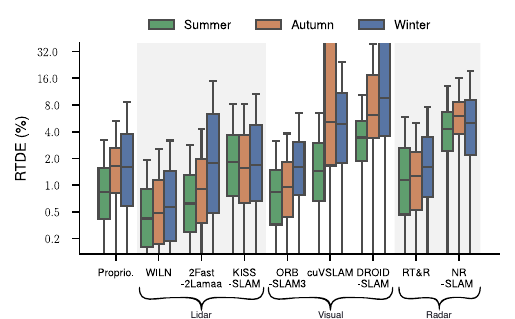}
    \caption{A comparison of \acf{RTDE} performance of proprioceptive, lidar, radar, and visual-based odometry algorithms across seasons.}
    \label{fig:results-boxplot-sensors-seasons}
\end{figure}

\rev{In this section, we will investigate how are the different methods and sensor modalities influenced by seasonal changes.}
To refine the analysis, we reduce the evaluation window size to \qty{10}{\meter} to accurately correlate sensor readings with the observed error.
\rev{Consequently, we utilize the \ac{RTDE}, as it avoids the orientation ambiguity around the forward axis of motion that has an effect on shorter windows when using \ac{SARTE}.
At such a short window size, segments of a trajectory could not be reliably aligned due to local trajectory jitter and sensor noise.
However, this metric switch comes at the cost of only focusing on the path length difference.
We report the distribution of \ac{RTDE} for the nine evaluated methods across selected deployments and trajectories in summer, autumn, and winter in \autoref{fig:results-boxplot-sensors-seasons}.}
We specifically choose the \texttt{Yellow} and \texttt{Orange} trajectories for this analysis, as their lack of urban structures better highlights the natural seasonal variations.
The evaluated sequences origin from \qty{20}{\kilo\meter} of data split equally between summer, autumn and winter.
Spring is omitted as our deployments don't cover the season with enough data points.
Moreover, we follow the methodology of \citet{Cumming2009_inference} and consider the difference between two distributions to be \emph{significant} if the median of one falls outside the \ac{IQR} of the other.
\rev{
As expected, all evaluated sensor modalities exhibit some degree of sensitivity to seasonal changes, generally demonstrating lower error rates in summer and higher errors during winter.
The proprioceptive baseline yields a median error below \qty{0.9}{\percent} in summer, which increases to \qty{1.8}{\percent} in both autumn and winter, and a larger \ac{IQR} in the winter months.
In contrast, the lidar-based methods are the least affected by seasonal transitions, with WILN maintaining a median error of less than \qty{0.6}{\percent} across all three seasons.
This performance can be attributed to the dense point clouds generated by the 128-channel RoboSense lidar, which capture both the highly variable immediate proximity of the \ac{UGV} and the structurally stable tree canopies.
The other two lidar-based methods suffer from local jitter in the estimated trajectory, which shows as a higher traveled distance and lead to higher \ac{RTDE}.
Furthermore, contrary to \citet{Baril2022}, we did not observe a major influence of the uncertainty caused by snow-covered trees (\ie~the forest corridor effect) on position estimation performance.
Among the visual odometry techniques, ORB-SLAM3 shows similar performance in summer and autumn, whereas both cuVSLAM and DROID-SLAM worsen significantly from summer to the other two seasons.
Finally, the radar-based approaches display a less pronounced seasonal trend, as the summer medians for both \ac{RTR} and Navtech-Radar-SLAM remain within the \ac{IQR} of the winter distributions.
}
Given the low inter-season significance in lidar and radar results, we conclude that ranging sensors-based localization and mapping is robust to seasonal changes, but that visual-based methods are sensitive to seasonal variations.
Therefore, the next section investigates the relation between localization error and image feature distributions across seasons.

\subsection{SEASONS AS A STRESSOR ON VISUAL FEATURE DETECTION}
\label{section:visual-feature-detection}

\begin{figure}[tb]
   \centering
   \includegraphics[width=\linewidth]{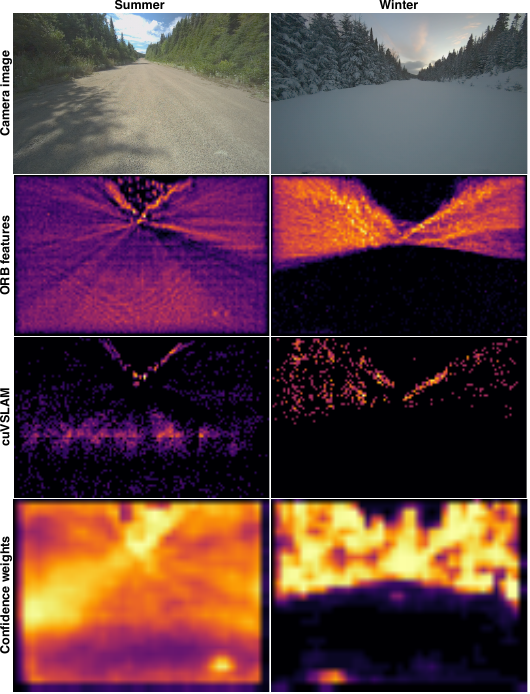}
   \caption{
   Effect of seasonal changes on ORB and cuVSLAM feature detection and DROID-SLAM network confidence.
   The top row shows ZED~X camera images taken in the \texttt{Orange} trajectory in summer and winter.
   \rev{The second and third rows displays heatmaps of the accumulated ORB and cuVSLAM features.}
   Bottom row shows a heatmap of accumulated DROID-SLAM confidence weights.
   The data originates from a ten-second window preceding the top row image capture.
   The color-maps are normalized for each individual image.
   }
   \label{fig:results-features-weights}
\end{figure}

For visual feature detection, we focus our study on the number and distribution of detected features in the left lens of the ZED~X camera image, as well as on the confidence weights of DROID-SLAM's update module.
\citet{Paton2017_expanding} already showed that in winter, detected \ac{SURF} tend to cluster on the horizon line.
Pixels around the horizon line, in the upper half of the image, have larger depth uncertainty and therefore reduce the accuracy of the translation estimation, directly affecting the \acf{RTDE}.
\citet{Paton2017_expanding} performed their experiments in open areas, where the horizon line corresponded to distant trees and other objects.
In our case, pixels in the upper half of the image mostly constitute the forest corridor (\ie~an invariable line of coniferous trees on the left and right side of the road).
An example of such a forest corridor is presented in \autoref{fig:results-features-weights}.
The figure contains an example image of the same location on the \texttt{Yellow} trajectory, recorded in summer (August 20th) and winter (January 29th) 2025.
The first row depicts the raw images originating from the left lens of the ZED~X camera, in summer and winter.
A wide dynamic range can be noticed in summer, with a significant exposure difference between highlights and shadows.
In winter, the intact snow in front of the \ac{UGV} appears flat and contains a few features.
The second row shows heatmaps of accumulated ORB~features over a ten-second window preceding the capture of the camera image in the first row.
While the feature density is relatively uniform in summer, in winter the features migrate to the upper half of the image.
\rev{
Similarly, the third row shows heatmap of keypoints detected by cuVSLAM based on the Shi-Tomasi metric.}
Finally, the last row shows the accumulated confidence weights of the \ac{BA} update module used by DROID-SLAM.
The confidence weights, between the left and right camera pairs, are displayed as a heatmap.
Although the update module's resolution is significantly lower compared to the raw image (\ie~1900$\times$1200 as opposed to 70$\times$43 pixels), we remark the network's lower confidence in the lower part of the image in winter.
Moreover, these confidence weights are only generated for key-frames, which are triggered by sufficient average optical flow relative to the previous key-frame.
Over the ten-second window, the network extracted 25 key-frames in summer but only a single key-frame in winter.
\rev{
\autoref{tab:results-features-distribution} shows statistics on the feature and confidence weights distribution between the upper and lower half of the image.
}

\begin{figure}[htpb]
    \centering
    \includegraphics[width=\linewidth]{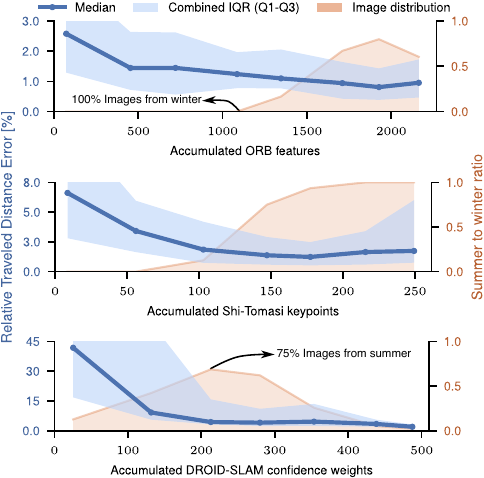}
    \caption{
    \rev{
    Relation between \acf{RTDE} and the distribution of ORB and cuVSLAM features and DROID-SLAM weights across seasons.
    \acs{IQR} stands for \acl{IQR}.
    Lower number of detected features leads to higher error values and correlates to images originating in winter months.
    Note that the scale of \ac{RTDE} differs between the three graphs.}
    }
    \label{fig:results-features-error}
\end{figure}

\rev{We extend the analysis of ORB feature, Shi-Tomasi keypoints and DROID-SLAM confidence weights migration in \autoref{fig:results-features-error}, where we inspect the effects of the migration on the reported \ac{RTDE}.}
Here, we only focus on the lower half of the image, as features in this part have the highest effect on translational accuracy~\cite{Paton2017_expanding}.
\rev{The figure shows the distribution of the \ac{RTDE}, as the median, together with first and third quartiles, as a function of the accumulated number of ORB and cuVSLAM features and DROID-SLAM confidence weights.}
To compute the errors, we take three summer and three winter recordings for the two longest trajectories: \texttt{Orange} and \texttt{Yellow}.
\rev{Note that DROID-SLAM reports less data points in each window $\Delta$ compared to the two other methods.}
In each of these windows, we compute and accumulate the number of features in the left lens ZED~X camera image, as well as the sum of confidence weights between the left and right camera images.
We then cluster these values based on similar features or weight counts and compute the ratio of how many data points come from summer and winter deployments, where 1.0~corresponds to all data points coming from summer and 0.0~all originating in winter.
We observe that a low number of features and a low confidence weight sum corresponds to higher error values.
There are always at least 1000~detected ORB features in the lower half of the image in summer, corresponding to errors under \qty{1.8}{\percent}.
The error is lowered under \qty{1}{\percent} as the number of detected features gets close to 2000, which we observe \qty{80}{\percent} of the time in summer.
However, the error, and the \ac{IQR} grows steeply when the number of features is under 500.
\rev{Similarly for cuVSLAM, the \ac{RTDE} exceeds \qty{7}{\percent} when processing exclusively winter images, which average fewer than 50 detected features.
The error decreases to under \qty{1}{\percent} when the method detects more than 175 features per image on average, a condition observed when the data consists almost entirely of summer images.}
Finally, in the case of DROID-SLAM, the error values reach over \qty{30}{\percent} in winter, where the sum of weights in the bottom part of the image is under 100.
The error drops to \qty{5}{\percent} as the ratio between summer and winter data reaches 0.6 or more, and the sum of weights in the lower part of the image climbs over 200.
Interestingly, around a weight sum of 220, the season ratio begins to decrease, indicating an increasing proportion of winter data points.
Despite this seasonal shift, the error remains low because the robot's translation is well constrained by the high confidence weights in the lower portion of the image.
These high confidence winter images primarily originate from sequences where the robot faces textured snowbanks, together with the \ac{UGV} traveling on plowed roads and from images capturing tracks left by preceding vehicles.
\rev{\autoref{tab:results-features-distribution} contains a summary of the per-bin median \ac{RTDE} values for the three evaluated methods.}

In general, our analysis indicates that the flat, featureless snow surface is the primarily source of the translational drift for \rev{all three evaluated visual-based methods}.
Conversely, these seasonal effects are less significant on plowed roads, where the snow layer does not appear as uniform.
To isolate the impact of high-contrast edges in deep snow, we perform limited experiments using the rear-mounted Basler camera.
These experiments show that the decay of detected ORB features is significantly less pronounced in images captured in the rear view.
We attribute this difference to the fact that the \ac{UGV} modifies the terrain in traverses and leaves tracks behind, which can serve as strong visual features in the images.
However, while these local features assist in maintaining odometry, they are season-dependent.
This leads to a more complex challenge: long-term localization across seasons.

\begin{table}[ht]
\centering
\caption{\rev{Effects of seasonal changes on visual-based methods.
\emph{Bottom half} denotes the percentage of features or confidence weights in the lower half of the image.
\acs{RTDE} limits report the range of per-bin median \ac{RTDE}, computed over bins where at least \qty{75}{\percent} of frames come from the respective season.
}}
\label{tab:results-features-distribution}
\begin{tabularx}{\columnwidth}{@{} X l r r @{}}
\toprule
\textbf{Method} & \textbf{Metric} & \textbf{Summer} & \textbf{Winter} \\
\midrule
ORB-SLAM3  & Bottom half & \qty{56}{\percent} & \qty{1.8}{\percent} \\
 (ORB features)         & \acs{RTDE}$_{\min}$ & \qty{0.74}{\percent} & \qty{1.31}{\percent} \\
                        & \acs{RTDE}$_{\max}$ & \qty{0.99}{\percent} & \qty{2.47}{\percent} \\
\midrule
cuVSLAM                         & Bottom half & \qty{74.5}{\percent} & \qty{0.8}{\percent} \\
(Shi-Tomasi keypoints)          & \acs{RTDE}$_{\min}$ & \qty{1.30}{\percent} & \qty{1.90}{\percent} \\
                                & \acs{RTDE}$_{\max}$ & \qty{1.85}{\percent} & \qty{7.35}{\percent} \\
\midrule
DROID-SLAM                      & Bottom half & \qty{41}{\percent} & \qty{2.9}{\percent} \\
(confidence weights)            & \acs{RTDE}$_{\min}$ & \qty{3.79}{\percent} & \qty{2.24}{\percent} \\
                                & \acs{RTDE}$_{\max}$ & \qty{3.79}{\percent} & \qty{43.66}{\percent} \\
\bottomrule
\end{tabularx}
\end{table}

\subsection{SEASONS AS A STRESSOR ON LOCALIZATION IN A PRIOR MAP}
\label{section:seasons-localization}

\begin{figure*}[t]
    \centering
    \includegraphics[width=\linewidth]{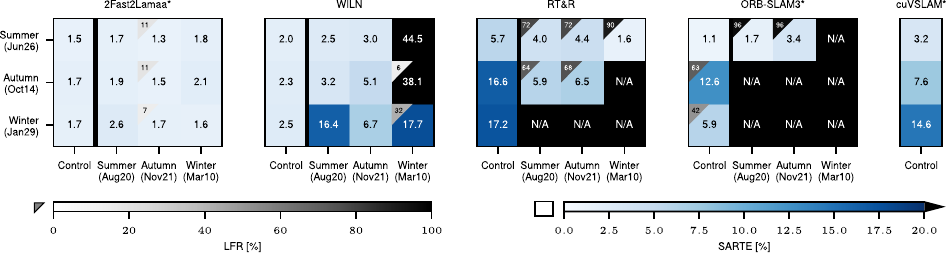}
    \caption{Evaluation of cross-seasonal localization performance on combined data from \texttt{Orange} and \texttt{Yellow} trajectories.
    Light cells mean better performance.
    Each row indicates the season and month used to construct the prior offline map.
    \rev{The Control columns shows localization performance on the same data as the mapping sequence.
    Each remaining column denotes the season and month of the active localization run.}
    The \acf{LFR}, displayed in the top triangle of each cell, is calculated relative to the full trajectory duration.
    \rev{We omit the ratio when the localization failed within the first \qty{100}{\meter} (\ie~black cells with N/A) without recovery, or when the full trajectory was completed successfully.
    \acs{SARTE} stands for \acl{SARTE}.
    Because of its architectural limitation, cuVSLAM was unable to successfully initialize outside of the three control sequences.
    }}
    \label{fig:results-cross-season-localization}
\end{figure*}

Long-term autonomy tasks, such as off-road environmental monitoring, require robust cross-seasonal localization.
We study this specific localization challenge in this final results section.
For this evaluation, we only consider the solutions that support localization along a path, as presented in \autoref{tab:methods} in the \emph{Loc. mode} column.
The five discussed methods are: WILN, 2Fast-2Lamaa, \ac{RTR}, ORB-SLAM3 and cuVSLAM.
Given a prior environment representation computed offline (\ie~a map), we evaluate the localization performance of the five methods in this prior map.
\rev{
The localization query runs were executed inside a containerized environment, and conducted on a high-performance workstation, featuring an Intel Core Ultra~9~285K~CPU, 192~GB of RAM and running Ubuntu~22.
Throughout the text, we use the terms query and localization interchangeably, as we do with offline and mapping.}
The starting point of the \rev{query} run is within one meter of the origin of the mapping run, and the localization trajectory follows the same waypoints, although it is not identical to the mapping run.
The \ac{UGV} operates within a \qty{10}{\meter} wide corridor on forest roads, maintaining a consistent direction of travel across both the mapping and localization runs.
To study the effect of seasonal variations on localization quality, we select three deployments as the source of our offline maps: June 26th (summer), October 14th (autumn), and January 29th (winter).
We report the results averaged over the \texttt{Yellow} and \texttt{Orange} trajectories, which total over \qty{4}{\kilo\meter}.
Similarly to the previous results sections, we selected these trajectories for their seasonal changes and lack of static urban landmarks.
\rev{
We first perform a control self-localization experiments, where the offline and query runs are identical.
Additionally, we also evaluate against three distinct deployments from the same seasons: August 20th (summer), November 21st (autumn), and March 10th (winter).}
\autoref{fig:results-cross-season-localization} depicts the resulting \acf{SARTE} together with the \acf{LFR}.
In the result matrices displayed in the figure, each row corresponds to the mapping (offline) month, and each column represents the localization (query) run.
Therefore, the diagonal illustrates localization within the same season, \rev{while the first column indicates the control experiment.}
Importantly, there are no persistent updates between the maps across runs, meaning methods cannot incrementally update the map from summer to autumn and then use it in winter.
This design choice intentionally targets the worst-case robustness of the localization methods by preventing them from leveraging incremental environmental changes.

\rev{
Relying on local maps, 2Fast-2Lamaa* demonstrated the best cross-seasonal localization performance among the tested methods.
Its \ac{SARTE} values closely aligned with the averaged odometry results previously presented in \autoref{fig:results-subarctic-performance}.
Although two specific winter mapping and localization scenarios exceeded a \qty{2}{\percent} error margin, the method generally remained highly robust against seasonal changes.
The localized failures observed in the November 21st sequence are likely an implementation error, as they not correspond to any notable disturbance in the query data.
WILN successfully completed the vast majority of its runs entirely in localization mode.
The exceptions included the autumn-winter combination, where it completed \qty{94}{\percent} of the trajectories, and the inter-winter pair, which resulted in a \qty{32}{\percent} \ac{LFR}.
In both of these instances, the method subsequently crashed and left the remainder of the trajectory unfinished.
The reported error for WILN peaked sharply between winter months and other seasons, reaching \qty{44.5}{\percent} when localizing inside a summer map using winter query data.
The results suggest that the source of the higher errors primarily originates from the presence of snow features in the sensor data rather than inconsistencies inside the prior map.
Similar to 2Fast-2Lamaa*, \ac{RTR} also utilizes local maps for its localization process.
Upon experiencing a localization failure, this method switches to odometry and briefly attempts to re-localize before permanently defaulting to dead reckoning for the remainder of the run.
While this fallback behavior suits its intended purpose as a \acl{TaR} solution, the overall low localization rates question the utility of the resource-intensive localization module.
Furthermore, \ac{RTR} systematically failed to localize at the beginning of the trajectory whenever winter data was introduced, whether used as a mapping or query sequence.
The only exception to this early failure pattern was the summer-to-winter combination, where localization only collapsed after completing \qty{10}{\percent} of the trajectory.
Finally, the observed error values for the control sequences originated from the \texttt{Orange} trajectories, mirroring the high error rates previously noted in \autoref{fig:results-subarctic-performance}.
The two visual methods were allowed a 100 frames initial window to localize, otherwise we declared a failure.
Under these constraints, ORB-SLAM3 proved extremely fragile when subject to significant scene changes.
Although the method could initially localize, it consistently lost track within the first \qty{100}{\meter}, making it impossible to compute a valid \ac{SARTE}.
It was also necessary to relax the real-time constraints applied to the other methods, as ORB-SLAM3's place recognition module could not operate in real time.
Within the control runs, the October 14th failure occurred during a sharp turn that suddenly introduced sun flares in the camera image.
Similarly, the January 29th control sequence failed when the robot transitioned from a road to a snowbank,
Unlike ORB-SLAM3*, cuVSLAM* does not perform global place recognition and instead relies on an initial local search.
Because of its architectural limitation, it was unable to initialize successfully when evaluated outside of the three control sequences.
Within those successful control sequences, the highest recorded error occurred during the winter runs, aligning with the findings of the previous sections.
}

The presented results show that cross-seasonal localization in subarctic environments remains a significant challenge for most tested methods, even when leveraging high-quality offline environment reconstructions.
\rev{Moreover, the high \ac{LFR} indicates that relying on a prior map can introduce additional system fragility.}
Specifically, radar is highly sensitive to the initial scan-to-map alignment, as the environment perceived by a 2D sensor changes drastically across seasons.
\rev{The visual-based ORB-SLAM3 completed only a single control trajectory entirely in localization mode.
It exhibited high sensitivity to cross-seasonal scene variations and proved fragile when consecutive frames lacked sufficient visual features, such as when climbing a snowbank.
Although the method is theoretically resilient, featuring a place-recognition backend designed to recover the robot's location, the uniform appearance of our sequence's forest roads provided too few distinct landmarks for successful relocalization.
Despite lacking long-term map management capabilities, WILN remained surprisingly stable even with reference maps that were several months old.
Finally, 2Fast-2Lamaa* consistently delivered strong performance across all tested sequences and seasons.
These contrasting outcomes highlight the broader challenge of achieving reliable long-term autonomy in unstructured subarctic terrain. The results suggest that different architectural choices heavily influence a system's tolerance to seasonal shifts.
Further investigation is required to generalize these conclusions across broader sensor modalities or method categories.
}

\section{CHALLENGES AND LESSONS LEARNED}

In this section, we use the experiments described in \autoref{section:experiments} and findings presented in \autoref{section:results} to draw the lessons learned from a year-long deployment of an \ac{UGV} in a boreal forest.
The total field time consisted of ten individual field days supplemented by five longer deployments, each spanning a full work week, for a total of 35~days.
We highlight remaining challenges for robot operations in subarctic regions and report on a field trial of multi-seasonal \ac{TaR}, executed in March 2025.
We conclude with a note on trajectory evaluation at scale in off-road conditions.

\subsection{ADDITIONAL PERTURBATIONS}

\begin{figure*}[htb]
    \centering
    \includegraphics[width=\linewidth]{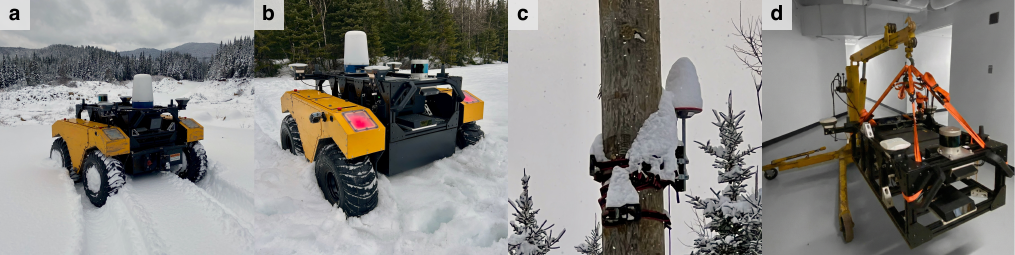}
    \caption{Challenges and lessons learned encountered during our year-long field campaign:
    (a, b) Mobility in snow is related not only to its depth, but also to other properties.
    (a) Shows the Warthog \ac{UGV} in the end of November, easily navigating over \qty{40}{\centi\meter} of fresh snow.
    In April (b), the robot's wheels were freely spinning as the wet spring snow didn't provide enough friction for the differential-drive robot to move forward.
    (c) Any permanently installed equipment, such as the depicted \ac{GNSS} antenna, must be cleaned regularly or equipped with heating elements to remove snow cover, which effectively blocks signals.
    (d) Performing a 6-\ac{DoF} manipulation with a half-a-tonne robot for extrinsic calibration requires special equipment.
    We used an industrial crane to excite all axes of the \ac{IMU}.
    }
    \label{fig:lessons-learned}
\end{figure*}

While our main analysis focused on the impact of the stresses on odometry, localization, and \ac{SLAM} algorithms, we identify multiple physical disturbances that we judge critical for any future deployment of autonomous \acp{UGV} in remote subarctic regions.
Notably, collisions with objects are common in off-trail environment.
During our recordings, we experienced 18 small collisions, mostly with boulders and small trees.
Moreover, the \acp{UGV} also suffered of two collisions with man-made structures including a gate and a sign post that imposed hard-stop on its course.
Although we did not observe any direct impact of these events on the odometry and localization performance, they should be taken into consideration when designing a vehicle for off-trail operation in subarctic region.
Consequently, fragile equipment, such as sensors, should be positioned to be protected against common collisions.
Similarly, we recorded 62 partial immobilizations in snow where insistent back-and-forth motions were required to resume the trajectory course.
The majority of immobilizations occurred on uphill trajectories over the same deployment in January, that saw a high level of fresh snow.
Despite being equipped with tracks, our Warthog \ac{UGV} easily becomes immobilized when the snow depth reaches the robot's main chassis.
In such a situation, the robot finds itself effectively parallelized with its motors spinning freely as either the wheels or tracks have no contact with the ground surface.
Additionally, the platform's mobility is challenging to assess only based on the snow depth.
In \autoref{fig:lessons-learned}-\emph{a}, we show the \ac{UGV} traversing through \qty{40}{\centi\meter} of fresh November snow without issues.
However, despite similar levels of snow, the robot struggled to move as the wheels spun freely, illustrated in \autoref{fig:lessons-learned}-\emph{b}, captured in mid-April.
The different behavior originates from the distinct snow properties.
While the fresh November snow had high cohesion and compacting it under the vehicle let to a sufficient traction, the wet April snow had a higher density and lower cohesion, leading to the \ac{UGV} removing material under its wheels, before getting stuck.
Snow also accumulates on all static equipment, including our static reference \ac{GNSS} antenna and other sensors, requiring regular cleaning for optimal performance \autoref{fig:lessons-learned}-\emph{c}.
Despite the lower received amount of sunlight, vegetation in subarctic regions can grow rapidly between months.
In our case, the \texttt{Green} trajectory, fully sheared in winter, was rapidly reclaimed by bush in the summer months, reaching up to \qty{2}{\meter} in height.
Bushes and tree canopy can hinder \ac{GNSS} coverage, especially when using single-band receivers.
We compare the number of \emph{fixed} \ac{PPK} points in \ac{GNSS} data for two types of receivers, mounted on the \ac{UGV}.
The \texttt{Green} trajectory saw the largest difference in the ratio of \emph{fixed} to \emph{float} and \emph{single} points, with single-band reporting \qty{56}{\percent} of \emph{fixed} solutions, compared to \qty{75}{\percent} for the multi-band receivers.
In the other trajectories, the ratios of fixed points are \qty{68}{\percent} and \qty{95}{\percent}, for single-band and multi-band devices, respectively.
These values are coherent with the number of observed satellites, where the multi-band receivers registered on average more than twice as many satellites as single-band.

As for other stresses not mentioned in \autoref{section:results}, subarctic regions exhibit significant thermal variability on both seasonal and short-term timescales.
Seasonal temperatures during our one year-long recording ranged from \qty{-34}{\degreeCelsius} in winter to \qty{25}{\degreeCelsius} in summer.
Notably, we saw the most extreme variation on May 28th, with temperatures ranging from \qty{-1}{\degreeCelsius} to \qty{25}{\degreeCelsius}, resulting in a \qty{26}{\degreeCelsius} temperature difference in a single day.
Cold weather, combined with the increased rolling resistance of tracks, \rev{significantly affects} the \ac{UGV}'s power autonomy in winter.
Our data show that the power consumption is approximately twice as high on tracks in winter, compared to wheels in summer months.
A different temperature range, combined with higher vibrations compared to the typical operating conditions of the \ac{UGV}, led to a battery module failure and a preliminary deployment termination in three cases.
The only experienced sensor failure caused by extreme temperatures, affecting the Basler camera, was recorded on the coldest day of the field trial in January, when temperatures were as low as \qty{-25}{\degreeCelsius}.
The extreme conditions compromised the auto-exposure system, leading to a continuous stream of overexposed images.

While autonomous cars are assumed to move on flat ground and therefore don't need to estimate the pitch and roll, an \ac{UGV} operating in subarctic environments must be able to deal with high attitude variations.
These variations specifically affect the Navtech radar, resulting in an increased number of ground strikes.
A deployment to a remote area makes it unfeasible to perform extrinsic calibration before each deployment, as the transport and temperature difference could make the calibration purposeless.
Furthermore, the use of standard solutions, such as the popular library Kalibr~\citep{Furgale2013}, proves to be challenging when attempting to perturb all 6-\ac{DoF} of our robotic platform for \ac{IMU}-camera extrinsic calibration.
While a typical road vehicle does not need to go through this type of calibration, as it is expected to operate on flat ground, a platform operating in subarctic environments should be able to cover a wider range of roll and pitch orientations.
We solved this issue by using an industrial crane with a lift capacity of over \qty{900}{\kilo\gram}, depicted in \autoref{fig:lessons-learned}-\emph{d}.
Beyond the perturbations documented during our 35~days in the field, further operational challenges are likely to emerge as subarctic autonomous deployments become more frequent.

\subsection{TEACH AND REPEAT ACROSS SEASONAL VARIATIONS}

Although \acf{TaR} is a well-established protocol for autonomous navigation, the ability of a robot to perform repeats across seasons \rev{using range sensors} has been little assessed.
\rev{While the work of \citet{Baril2022} includes a study on \ac{LTR} between winter and autumn months, the ability of a robot to repeat a trajectory taught in a snow deprived environment in a context of additional snow accumulation has not been explored yet.}
\rev{Furthermore, to the best of our knowledge, \acf{RTR} has not yet been tested against significant ground terrain alterations}.
In March 2025, we attempted cross-seasonal \ac{TaR} with a 4-day, 44-day, and 113-day time span between the teach and the repeat phases.
We chose the \texttt{Blue} trajectory for our tests, as its starting position next to a building provides enough initial localization features, while at the same time, the trajectory presents an adequate challenge with a transition to a deeper snow area and features unplowed forest roads.
The repeats were performed on the same day using three different sets of teach data from previous deployments.
Each teach run was repeated with both \ac{RTR} and \ac{LTR}.
The experiment highlighted several challenges associated with either both or one of the benchmarked methods.

\begin{figure}[t]
    \centering
    \begin{overpic}[width=\linewidth]{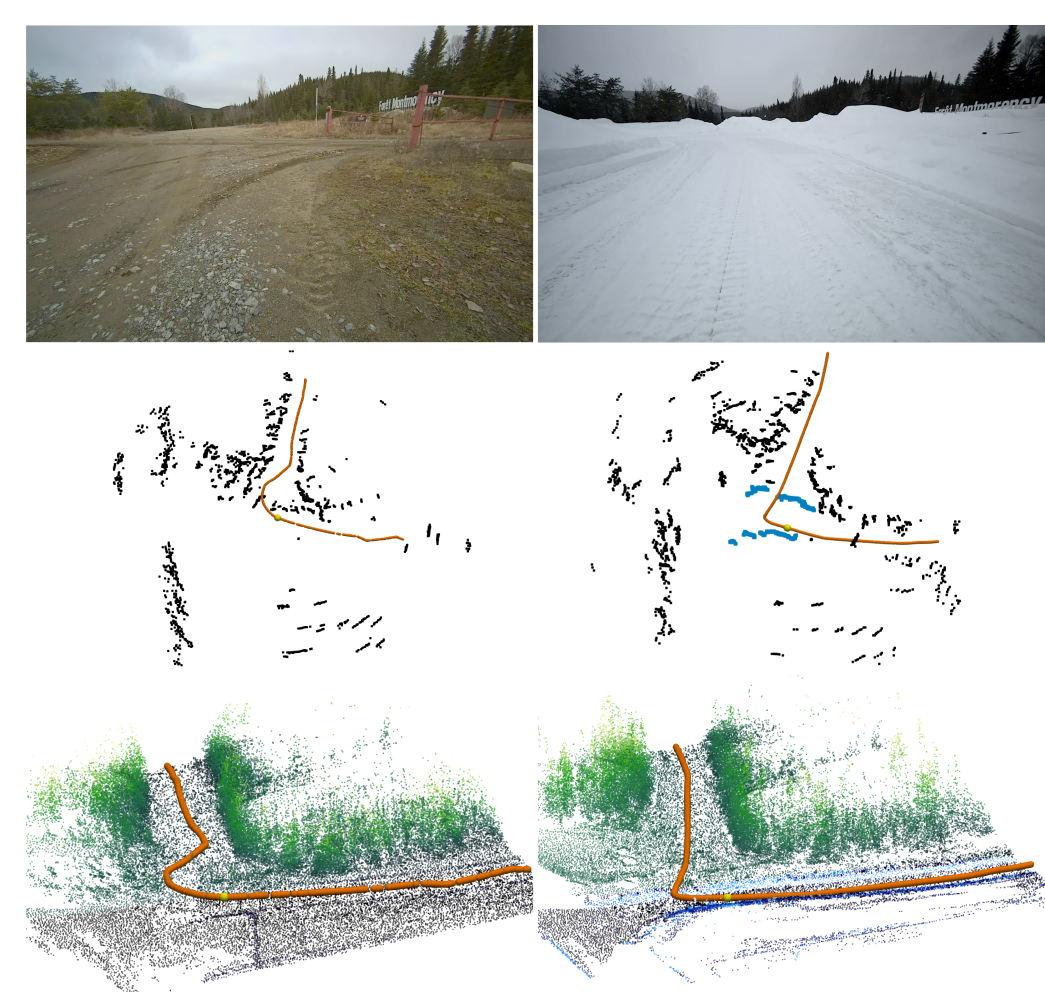}
        \put(0,0){\includegraphics[width=\linewidth]{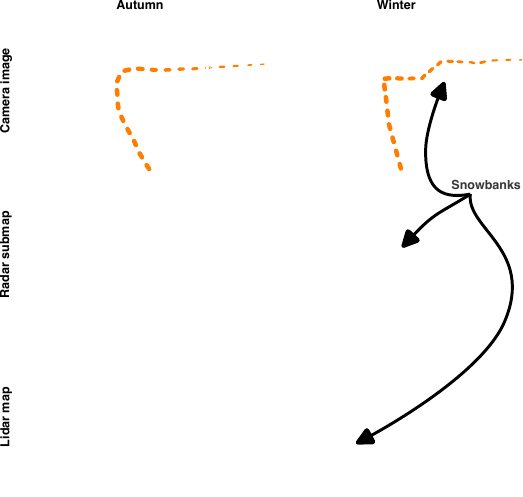}}
    \end{overpic}
    \caption{
    Sensor data comparison between seasons.
    First row shows images from the left lens of the ZED~X camera.
    Second row displays radar submaps with the robot's trajectory in orange and snowbanks highlighted in blue.
    The last row depicts a part of the global lidar point cloud, again with the robot's trajectory in orange.
    Green indicates height and vegetation, while the blue color in the winter column highlights snow accumulation around the road.
    Yellow spheres in the orange trajectories indicate the \ac{UGV}'s position where the corresponding image in the top row was taken. }
    \label{fig:gate}
\end{figure}

The main difference between lidar and radar localization is that radar is a two-dimensional sensor.
Therefore, \ac{RTR} relies on the assumption that the roll and pitch of the \ac{UGV} is similar between the teach and the repeat run to match the scans from both phases.
However, snow accumulation itself can modify the orientation of the robot at different points of the trajectory, especially when the robot travels on unplowed terrain.
We experience this phenomena especially during transitions between plowed and unplowed sections of the path, as a sudden pitch change occurs as the robot climbs over the snowbank, such as the one in \autoref{fig:gate}.
When the robot drives uphill, the sensing plane of the radar is also tilted, leading to ground strikes on one side and lack of features on the other side of the scan.
Although the \ac{LTR} system is immune to faults from attitude discrepancy due to its 3D~nature, it faced other challenges that made localization difficult.
Notably, when performing repeat from a 113-days-old teach run, the \ac{ICP}-based lidar localization failed in a plowed area where snowbanks height reached up to \qty{3}{\meter}.
We found that by filtering out all lidar returns under \qty{1.5}{\meter} above the sensor position (\ie~\qty{2.6}{\meter} above ground plane), the algorithm had enough features from tall objects and tree tops to localize in the environment.
However, this solution \rev{might not be transferable} to lidar sensors with narrower vertical field of view than the employed 128-beam RoboSense Ruby Plus.

A remaining challenge for both \ac{LTR} and \ac{RTR} is the controller's behavior when subject to snow-covered ground.
The motion of the robot deviates from the prediction of the motion model implemented in the controller when driving on snow, which results in poor prior given to the rest of the localization process.
Sharp turns proved to be the hardest to execute on snow, demanding more energy and track rotational speed compared to hard ground due to reduced traction.
The experiment highlighted the tight connection between localization and motion controllers, demonstrating that hindered motion usually also results in poor localization and, subsequently, higher path tracking errors.

\subsection{TRAJECTORY EVALUATION AT SCALE}

Current benchmarking of odometry and localization methods typically relies on standard error metrics, such as the \ac{ATE} or \ac{RTE}, to compare performance on public datasets.
While these benchmarks facilitate direct comparison between different methods, we argue that the conditions themselves should be more easily comparable.
As reported in \autoref{section:results-subarctic-performance-report}, the quality of odometry degrades significantly under subarctic conditions.
Given that datasets in extreme climate remain scarce and small in scale compared to their urban driving counterparts, there is a clear need for a more standardized characterization of the testing environment.
We propose the adoption of the Köppen-Geiger climate classification to provide context on the seasonal changes in the recorded data, as well as the public benchmarks.
This context would ease the comparison of results between different environments in year-round data recordings, and would help when discussing the adaptability of odometry, localization and mapping methods.

Additionally, standard error metrics are informative as long as the estimate and reference trajectories remain reasonably similar.
However, in meteorologically challenging environments or environments lacking distinct features, systems can experience catastrophic failures, which are, under certain conditions, hidden by the employed error metrics.
Although it is easy to count the number of failures due to crashing software, the transitions between high errors due to poor state estimation and complete failures remains unclear.
This lack of a common error metric makes the selection of the best solution difficult, as methods with low errors can also report high failure rates and vice-versa, as shown on the example of ORB-SLAM3 and DROID-SLAM in \autoref{section:results-loop-closure-subarctic}.

\rev{
Finally, acquiring unbiased and independent \acl{GT} trajectories is challenging, particularly for large-scale experiments.
While \ac{GNSS} combined with \ac{RTK} or \ac{PPK} can provide accurate position estimates, it lacks the orientation information required by standard error metrics.
A potential workaround is to employ an array of \ac{GNSS} receivers and estimate the orientation from their known geometry.
The deployed \ac{UGV} is equipped with three receivers featuring baselines of \qty{0.90}{\meter}, \qty{1.15}{\meter}, and \qty{0.79}{\meter} between antennas.
However, our experiments demonstrate that orientation estimated through point-to-point minimization is highly sensitive to noise in the \ac{GNSS} elevation measurements.
Additionally, the \ac{GNSS} observation uncertainty is frequently underestimated by the processing software.
Nevertheless, as this report shows, the reported error values remain substantial even without evaluating against full 6-\ac{DoF}~\ac{GT} trajectories, indicating that the primary challenges for autonomous navigation in subarctic environments currently originate from other sources.
}

\section{CONCLUSION}
\rev{
In this paper, we formulate a methodology for evaluating odometry, localization and mapping algorithms outside of the typical urban or laboratory conditions, and utilize this methodology to evaluate nine radar-, lidar- and visual-based methods.
The methods are evaluated on over \qty{64}{\kilo\meter} of recorded data, covering seasonal changes in subarctic environment over the course of twelve months.
We demonstrated that state-of-the-art methods suffer significant performance degradation under these conditions, whereas simple proprioceptive odometry exhibits surprisingly strong robustness, even in off-trail terrain.
Additionally, we analyze the effects of seasonal variations on the performance of lidar, radar, and camera-based methods.
We note that while the performance of range-sensing modalities does not significantly degrade, both feature- and deep-learning-based visual methods experience a shift of feature or weight distribution between seasons, directly impacting translational drift.
In the task of cross-season localization, the initial state estimation and subsequent tracking pose a challenge for state-of-the-art visual- and radar-based methods, while lidar methods strongly benefit from the wide vertical field of view of the employed sensor.}

\rev{
Applying our proposed time series taxonomy to these field trials highlighted the difficulty of attributing performance degradation.
While environmental stressors, such as seasonal changes, are the main focus, measured performance is inherently influenced by unmodeled internal factors, such as implementation quality.
To better isolate and contextualize these challenges, we encourage future field reports to include proprioceptive odometry as a standard baseline, report the Köppen climate classification of deployment sites to facilitate environmental comparisons, and explicitly detail failure rates for algorithms operating outside laboratory conditions.}

\rev{
While our evaluation highlights the capabilities of current algorithms, achieving robust long-term autonomy across seasonal variations in off-road subarctic environments still demands further investigation.
The high cross-season \acl{LFR} observed for both visual and radar methods requires further attention.
Future work will involve a dedicated study on place recognition in subarctic environments.
Furthermore, we intend to investigate error metrics better suited for large-scale field deployments.
Finally, we plan to expand our \acl{TaR} field tests by repeating taught trajectories across a more diverse range of seasonal combinations to further evaluate system reliability.}

\section*{APPENDIX}
\label{section:appendix}

\begin{figure*}
    \centering
    \includegraphics[width=\linewidth]{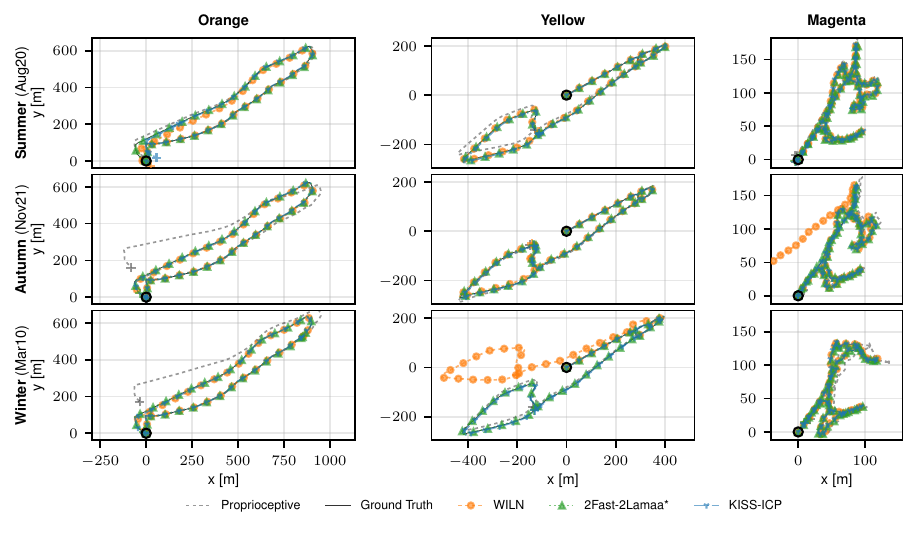}
    \caption{Qualitative results for lidar odometry methods and the Proprioceptive baseline.}
    \label{fig:qualitative-odom-lidar}
\end{figure*}

\begin{figure*}
    \centering
    \includegraphics[width=\linewidth]{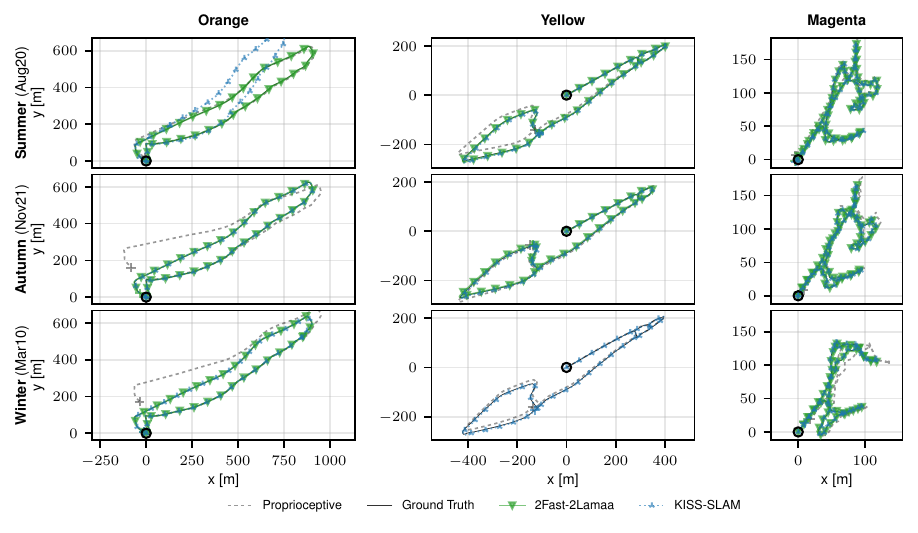}
    \caption{Qualitative results for lidar odometry methods with \acl{PGO} and the Proprioceptive baseline.}
    \label{fig:qualitative-slam-lidar}
\end{figure*}

\begin{figure*}
    \centering
    \includegraphics[width=\linewidth]{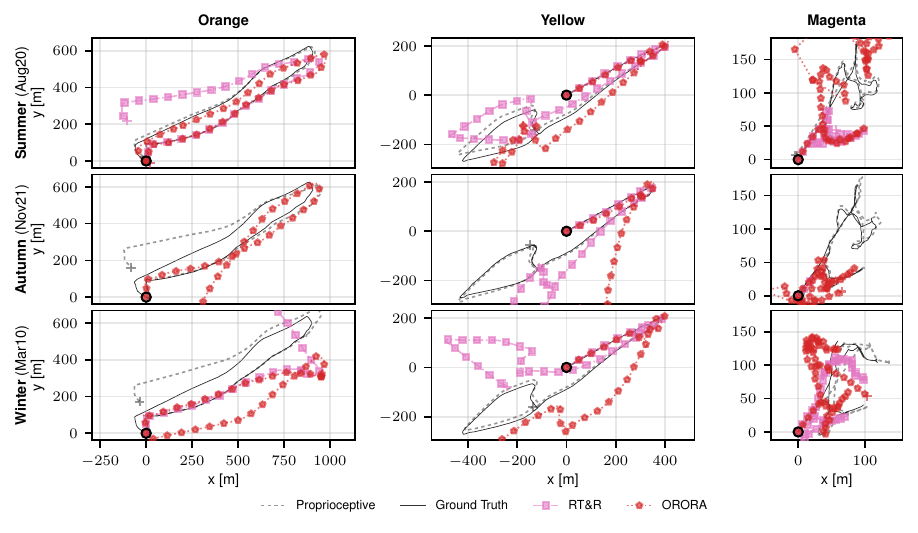}
    \caption{Qualitative results for radar odometry methods and the Proprioceptive baseline.}
    \label{fig:qualitative-odom-radar}
\end{figure*}

\begin{figure*}
    \centering
    \includegraphics[width=\linewidth]{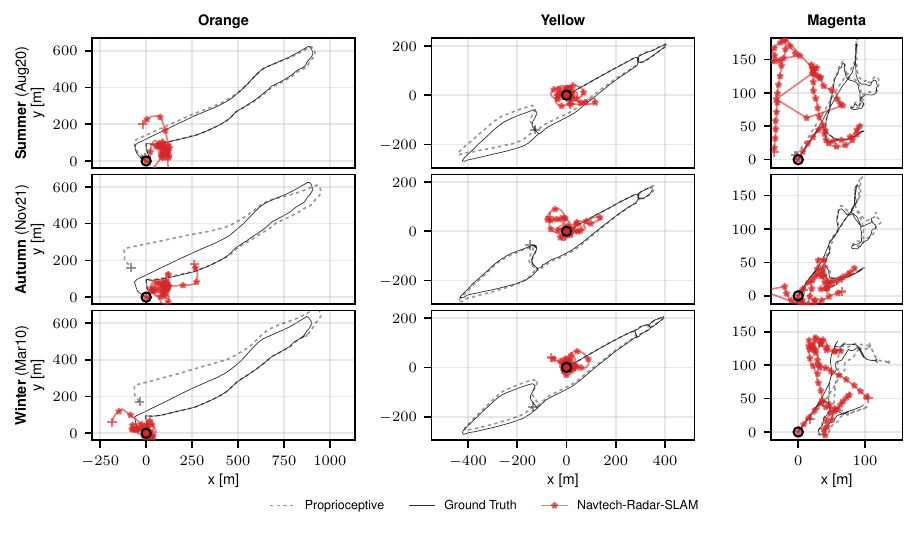}
    \caption{Qualitative results for lidar odometry methods with \acl{PGO}  and the Proprioceptive baseline.}
    \label{fig:qualitative-slam-radar}
\end{figure*}

\begin{figure*}
    \centering
    \includegraphics[width=\linewidth]{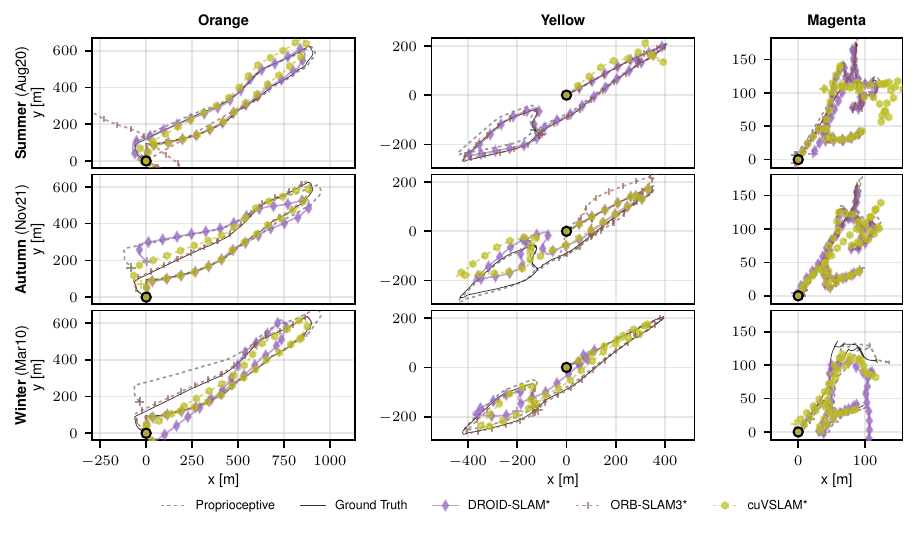}
    \caption{Qualitative results for visual odometry methods and the Proprioceptive baseline.}
    \label{fig:qualitative-odom-visual}
\end{figure*}

\begin{figure*}
    \centering
    \includegraphics[width=\linewidth]{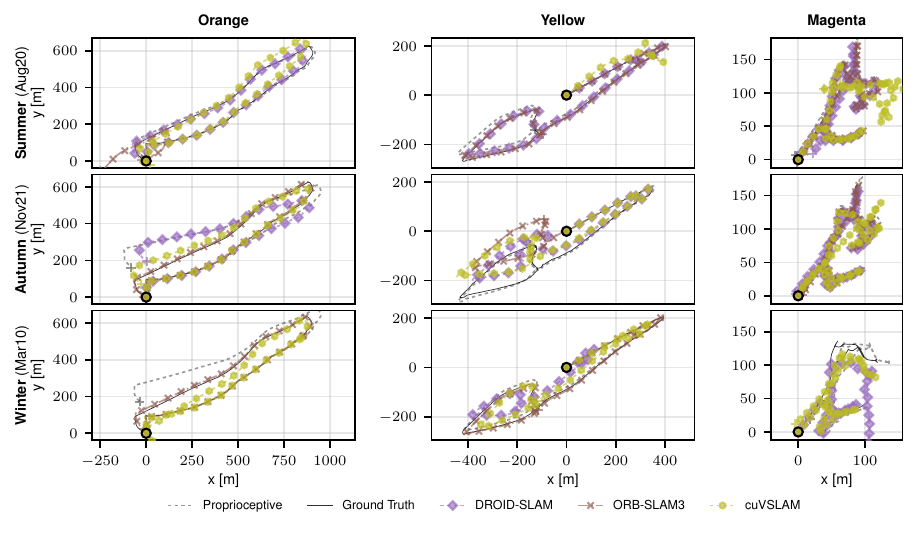}
    \caption{Qualitative results for visual odometry methods with \acl{PGO}  and the Proprioceptive baseline.}
    \label{fig:qualitative-slam-visual}
\end{figure*}

\rev{
In this section, we present selected qualitative results for the \texttt{Orange}, \texttt{Yellow}, and \texttt{Magenta} trajectories across three seasonal deployments.
The full results are available in a Docker-based viewer. 
\autoref{fig:qualitative-odom-lidar} and \autoref{fig:qualitative-slam-lidar} illustrate the lidar-based odometry and odometry with \acf{PGO} methods, respectively.
Notably, WILN diverges from the \acl{GT} trajectory due to snow banks during the winter \texttt{Yellow} deployment.
Furthermore, while \ac{PGO} improves the scale estimation from KISS-ICP to KISS-SLAM in the summer \texttt{Orange} trajectory, it introduces an orientation misalignment.
\autoref{fig:qualitative-odom-radar} shows the odometry trajectories for the two radar-based methods.
The November~21st trajectory for \ac{RTR} is omitted due to an initialization failure.
Additionally, \autoref{fig:qualitative-slam-radar} demonstrates the impact of false-positive loop closures on the trajectory quality of Navtech-Radar-SLAM.
\autoref{fig:qualitative-odom-visual} and \autoref{fig:qualitative-slam-visual} present the results for the three visual odometry methods, with and without \acf{PGO}.
ORB-SLAM3 exhibits the greatest benefit from \ac{PGO}.
In contrast, no difference is observed between the cuVSLAM and cuVSLAM* trajectories.
Finally, DROID-SLAM incorrectly estimates the scale of the \texttt{Yellow} trajectory during the autumn and winter deployments.
}

\section*{ACKNOWLEDGMENT}
We would like to thank the team at Forêt Montmorency for their assistance in organizing our deployments.
We thank Alexander Krawciw from the University of Toronto for his help with \acl{RTR} field trial and evaluation.
Our gratitude also goes to Vsevolod Hulchuk from the Czech Technical University in Prague, whose efforts helped bootstrap the development of our evaluation pipeline.

\printbibliography[title={\MakeUppercase{References}}]

\vfill\pagebreak

\end{document}

%% file: preamble.tex
\usepackage[
    backend=biber,
    style=ieee,
    sorting=none,
    natbib=true,
    doi=false,
    isbn=false,
    url=false,
    eprint=false,
    maxcitenames=1,
    mincitenames=1
]{biblatex}

\usepackage[colorlinks,bookmarksopen,bookmarksnumbered,citecolor=red,urlcolor=red]{hyperref}

\usepackage[english,noabbrev,nameinlink]{cleveref}

\usepackage[printonlyused]{acronym}

\usepackage{siunitx}
\DeclareSIUnit{\pp}{\%pt} 

\usepackage[all]{nowidow}

\usepackage[dvipsnames]{xcolor}

\usepackage{lipsum}

\usepackage{xspace} 
\newcommand{\ie}{i.e.,\xspace{}}
\newcommand{\eg}{e.g.,\xspace{}}

\usepackage{xurl}

\usepackage{epstopdf}

\usepackage{overpic}

\usepackage{import}

\usepackage{booktabs}

\usepackage{tabularx}
\usepackage{multirow, multicol}

\usepackage{amssymb,amsfonts,amsmath,amscd}

\usepackage{bm}

\usepackage{algorithm}
\usepackage{algpseudocode}

\usepackage{dirtree}

\usepackage[english]{datetime2}
\DTMsetdatestyle{ddmmyyyy}

\newcommand{\bbm}{\begin{bmatrix}}
\newcommand{\ebm}{\end{bmatrix}}

\newcommand{\rev}[1]{{#1}}

\acrodef{ICP}{Iterative Closest Point} 
\acrodef{MOCAP}{Motion Capture} 
\acrodef{SLAM}{Simultaneous Localization and Mapping}
\acrodef{VSLAM}{Visual SLAM} 
\acrodef{IMU}{Inertial Measurement Unit} 
\acrodef{GT}{Ground Truth} 
\acrodef{FoMo}{For\^{e}t Montmorency} 
\acrodef{SDK}{Software Development Kit}
\acrodef{GNSS}{Global Navigation Satellite System} 
\acrodef{UGV}{Uncrewed Ground Vehicle} 
\acrodef{UAV}{Uncrewed Aerial Vehicle} 
\acrodef{FMCW}{Frequency Modulated Continuous Wave}
\acrodef{PTP}[PTP]{IEEE1588 Precision Time Protocol}
\acrodef{NMEA}[NMEA]{National Marine Electronics Association}
\acrodef{PPK}[PPK]{Post Processed Kinematic}
\acrodef{RTK}[RTK]{Real Time Kinematic}
\acrodef{PPS}[PPS]{pulse-per-second}
\acrodef{DL}[DL]{Deep Learning}
\acrodef{GMSL2}[GMSL2]{Gigabit Multimedia Serial Link 2}
\acrodef{INS}{Inertial Navigation System}
\acrodef{AE}{Auto Exposure}
\acrodef{ROI}{Rectangle of Interest}
\acrodef{RANSAC}{Random Sample Consensus}
\acrodef{RTS}{Robotics Total Station}
\acrodef{BEV}{Bird's-eye view}
\acrodef{ETH}{Ethernet}
\acrodef{RINEX}{Receiver Independent Exchange Format}
\acrodef{CORS}{Continuously Operating Reference Station}
\acrodef{DoF}{Degree of Freedom}
\acrodefplural{DoF}[DoFs]{Degrees of Freedom}
\acrodef{RTR}[RT\&R]{Radar Teach and Repeat}
\acrodef{LTR}[LT\&R]{Lidar Teach and Repeat}
\acrodef{RGTR}[RT\&R]{Radar-Gyro Teach and Repeat}
\acrodef{TaR}[T\&R]{Teach and Repeat}
\acrodef{NLP}{Natural Language Processing}
\acrodef{FOV}{Field of View}
\acrodef{RPE}{Relative Pose Error}
\acrodef{IDD}{Ideal Differential Drive}
\acrodef{APE}{Absolute Pose Error}
\acrodef{ATE}{Absolute Trajectory Error}
\acrodef{RTE}{Relative Trajectory Error}
\acrodef{RE}{Relative Error}
\acrodef{PGO}{Pose Graph Optimization}
\acrodef{BA}{Bundle Adjustment}
\acrodef{SURF}{Speeded-up Robust Features}
\acrodef{IQR}{Inter-quartile Range}
\acrodef{SARTE}{Sequentially Aligned Relative Trajectory Error}
\acrodef{RTDE}{Relative Traveled Distance Error}
\acrodef{LFR}{Localization Failure Ratio}
\acrodef{CFAR}{Constant False Alarm Rate}
\acrodef{CORS}{Continuously Operating Reference Station}
\acrodef{GPR}{Ground Penetrating Radar}

\usepackage{fontawesome5}
\usepackage{xcolor}

\newcommand{\Checkmark}{\textcolor{Green}{\faCheck}}

\newcommand{\used}{\Checkmark}
\newcommand{\notused}{--}

\newcommand{\rot}[1]{%
    \rotatebox[origin=lb]{45}{\hspace{-1mm}\scriptsize\textbf{#1}}%
}

\makeatletter
\newcommand{\showfont}{%
  Font family: \f@family, %
  Font series: \f@series, %
  Font shape: \f@shape, %
  Font size: \f@size pt%
}
\makeatother

\usepackage{etoolbox}
\newcommand{\shortautoref}[2]{\hyperref[#1]{\tiny \ifstrempty{#2}{}{#2.}\ref*{#1}}}